\documentclass[lettersize,journal]{IEEEtran}
\usepackage{amsmath,amsfonts}
\usepackage{algorithmic}
\usepackage{algorithm}
\usepackage{array}
\usepackage[caption=false,font=normalsize,labelfont=sf,textfont=sf]{subfig}
\usepackage{textcomp}
\usepackage{stfloats}
\usepackage{url}
\usepackage{verbatim}
\usepackage{graphicx}
\usepackage{cite}
\usepackage{booktabs}
\usepackage{multirow}
\usepackage{color}
\usepackage[table]{xcolor}
\usepackage{hyperref}
\begin{document}

\title{A Low-rank Matching Attention based Cross-modal Feature Fusion Method for Conversational Emotion Recognition}

\author{Yuntao~Shou, Huan Liu$^*$, 
	Xiangyong Cao, Deyu Meng
	and~Bo Dong \\
	\thanks{$^*$ Corresponding Author: Huan Liu~(huanliu@xjtu.edu.cn)}
	\thanks{This work is supported by the National Nature Science Foundation of
	China (No. 62272374, No. 62192781, No. 62250009, No. 62137002, No. 62202367, No. 62272375), Natural
	Science Foundation of Shaanxi Province (No. 2024JC-JCQN-62), Project of
	China Knowledge Center for Engineering Science and Technology, Project of
	Chinese academy of engineering “The Online and Offline Mixed Educational
	Service System for ‘The Belt and Road’ Training in MOOC China”, and the
	K. C. Wong Education Foundation.}
	\IEEEcompsocitemizethanks{\IEEEcompsocthanksitem Y. Shou, X. Cao, and B. Dong are with School of Computer Science and Technology, the Ministry of Education Key Laboratory for Intelligent Networks and Network Security, and the Shaanxi Province Key Laboratory of Big Data Knowledge Engineering, Xi’an Jiaotong, Xi’an, Shaanxi 710049, China.
		(shouyuntao@stu.xjtu.edu.cn,
		caoxiangyong@mail.xjtu.edu.cn,
		dong.bo@mail.xjtu.edu.cn).}
		
		\IEEEcompsocitemizethanks{\IEEEcompsocthanksitem H. Liu are with School of Computer Science and Technology, Xi’an Jiaotong, Xi’an, Shaanxi 710049, China, and the BigKE Joint Innovation Center, SHMEEA, Shanghai, China. (huanliu@xjtu.edu.cn).}
		
		\IEEEcompsocitemizethanks{\IEEEcompsocthanksitem D. Meng is with School of Mathematics and Statistics, Xi’an Jiaotong University, Xi’an, China. (dymeng@mail.xjtu.edu.cn).}

}



\maketitle

\begin{abstract}
	Conversational emotion recognition (CER) is an important research topic in human-computer interactions. {Although recent advancements in transformer-based cross-modal fusion methods 
	have shown promise in CER tasks, they tend to overlook the crucial intra-modal and inter-modal emotional interaction or suffer from high computational 
	complexity. To address this, we introduce a novel and lightweight cross-modal 
	feature fusion method called Low-Rank Matching Attention Method (LMAM). LMAM 
	effectively captures contextual emotional semantic information in 
	conversations while mitigating the quadratic complexity issue caused by the 
	self-attention mechanism. Specifically, by setting a matching weight and calculating inter-modal features attention scores row by row, LMAM requires only one-third of the parameters of self-attention methods. We also employ the low-rank decomposition 
	method on the weights to further reduce the number of parameters in LMAM. As a result, LMAM offers a lightweight model while avoiding overfitting problems caused by a large number of parameters. Moreover, LMAM is able to fully exploit the intra-modal emotional contextual information within each modality and integrates complementary emotional semantic information across modalities by computing and fusing similarities of intra-modal and inter-modal features simultaneously. Experimental results verify the superiority of LMAM compared with other popular cross-modal fusion methods on the premise of being more lightweight. Also, LMAM can be embedded into any existing state-of-the-art CER methods in a plug-and-play manner, and can be applied to other multi-modal recognition tasks, e.g., session recommendation and humour detection, demonstrating its remarkable generalization ability.}
\end{abstract}

\begin{IEEEkeywords}
Attention, Cross Modal Feature Fusion, Low Rank Decomposition, Multimodal Emotion Recognition
\end{IEEEkeywords}

\section{Introduction}

{With the development of the multi-modal research field, Conversational Emotion Recognition (CER) that utilizes three modal data (i.e., video, audio and text) to identify the speaker’s emotional changes during the conversation has become a hot research topic \cite{mohammad2022ethics, geetha2024multimodal}.} Nowadays, CER has shown its promising performance in many practical social media scenarios. For example, in the field of intelligent recommendation, a recommendation system with emotional tendencies can recommend products that users are more interested in by identifying changes in consumers’ emotions. Therefore, it is of great importance to accurately identify the speaker’s emotional changes during the conversation.


\begin{figure}
	\centering
	\includegraphics[width=0.95\linewidth]{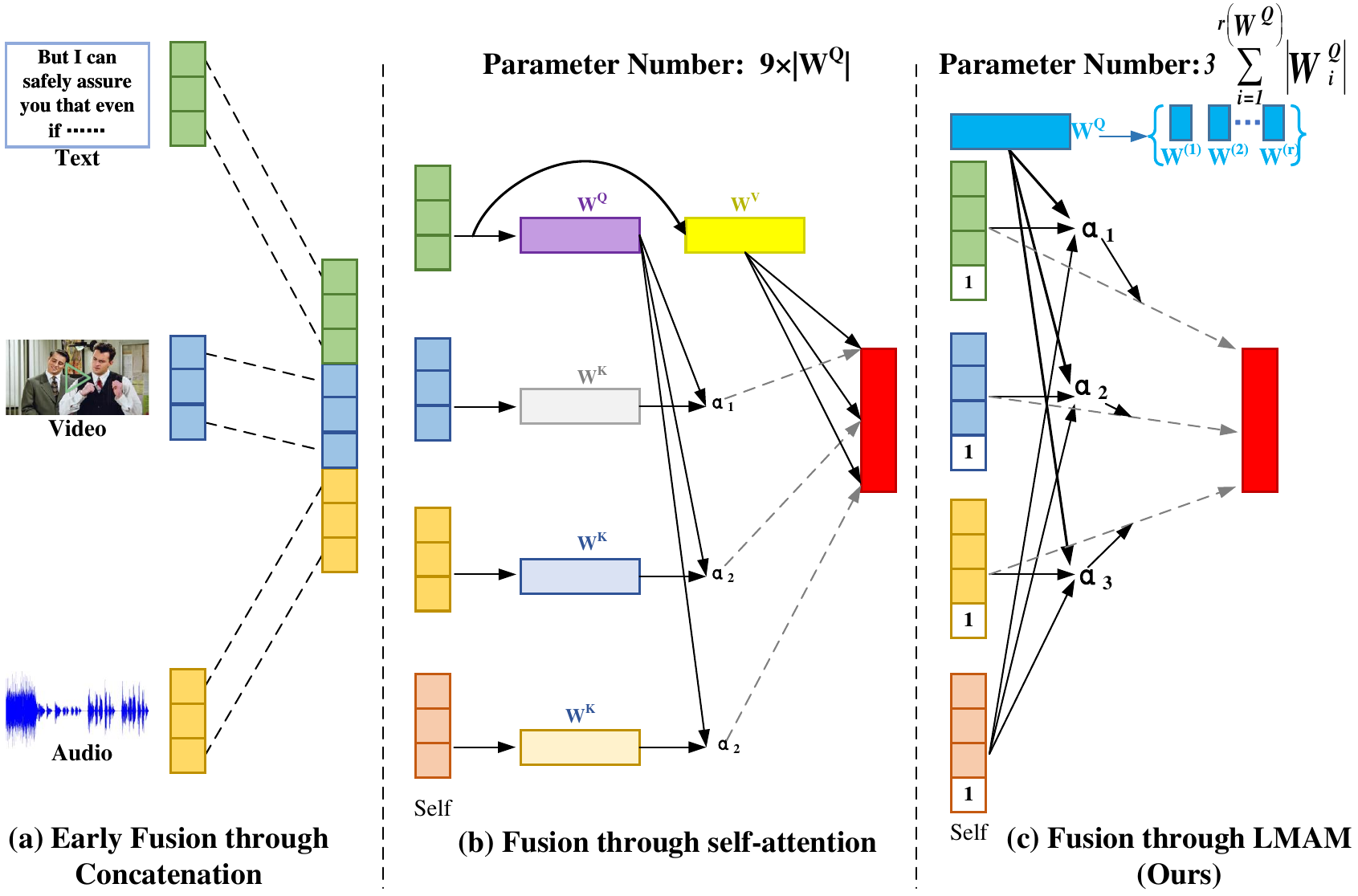}
	\caption{(a) Early fusion through multimodal concatenation. (b) Multimodal feature fusion through self-attention. $W^Q$, $W^K$, and $W^V$ {represent} the learnable parameters of the query, key, and value vectors respectively. The parameter number of {self-attention} is 9 times of the parameter number of $W^Q$. (c) Multimodal feature fusion through LMAM. The parameter number of our LMAM fusion method is less than 3 times of the parameter number of $W^Q$. $|W^{Q}_i|$ means the parameter number of the column vectors of $W_Q$ after low rank decomposition. $r(\cdot)$ indicates the rank of the matrix.}
	\label{figure1}
\end{figure}

{Many existing studies \cite{ren2021lr, shou2022conversational, tu2022exploration} have proven that multi-modal feature fusion plays an important role in CER tasks. For example, as shown in Fig. \ref{figure1}(a), previous work \cite{cambria2018benchmarking} achieved multi-modal feature fusion by simply concatenating multi-modal features. Although this method has utilized multimodal data to a certain extent, it often adopts a strategy of treating each modality equally without fully considering the difference in the amount of emotional information between different modalities. The above method has two limitations: 1) the modality with less information may introduce noise during the fusion process, weakening the performance of the overall model; 2) the dominant role of the modality with more information in emotion recognition is not fully utilized. In addition, existing methods cannot capture contextual emotional semantic information within and between modalities, which limits the performance of CER tasks.}

{To tackle the above problems, many Transformer-based methods (e.g., CTNet \cite{lian2021ctnet}, and EmoCaps \cite{li2022emocaps}, etc) have been proposed and have become the current mainstream multi-modal feature fusion methods in CER task. Specifically, as shown in Fig. \ref{figure1}(b), these Transformer-based multi-modal feature fusion methods mainly use the self-attention mechanism to calculate the attention scores of different modal features, and achieve the fusion of intra-modal and inter-modal contextual emotional semantic information based on attention scores. {Although the Transformer-based method can more accurately capture and fuse multimodal features in CER tasks, it also has two limitations. 1) This attention-based method has high computational complexity, especially when processing large-scale data, the demand for computing resources increases significantly. 2) As the model complexity increases, the risk of overfitting also increases accordingly, especially when the amount of training data is insufficient or the data diversity is limited. This may lead to a decrease in the generalization ability of the model and make it difficult to maintain stable performance in practical applications.}


{To solve these problems, existing research proposes low-rank decomposition methods \cite{kolda2009tensor} to reduce the number of parameters. For example, Zhu et al. \cite{zhu2020multimodal} proposed a Low-rank Tensor Multi-modal Fusion method to achieve multi-modal data fusion through a low-rank tensor outer product and attention mechanism. Jin et al. \cite{jin-etal-2020-dual} proposed a Fine-grained Temporal Low-rank Multi-modal Fusion (FT-LMF) method to fuse multi-modal information in time and space through low-rank tensor outer product. Sahay et al. \cite{sahay2020low} proposed Low-rank fusion based transformers to extract potential interactive information in multi-modal information through low-rank tensor outer product and cross-modal Transformer. {Although the above method achieves the fusion of multimodal information through low-rank tensor outer product, this approach has significant limitations. Specifically, this fusion method lacks fine-grained processing capabilities when capturing the potential relationship information between discourses. Although the low-rank tensor outer product method can reduce the computational complexity to a certain extent, due to its inherent structural constraints, it cannot fully express the complex interactions between modalities, resulting in inaccurate capture of potential relationship information.} 

{Different from the aforementioned CER methods, we thoroughly leverage the respective advantages of low-rank decomposition and Transformer techniques. Specifically, as shown in Fig. \ref{figure1}(c), we directly introduce a low-rank weight $\boldsymbol{W}$ into the self-attention, and then perform row-by-row attention calculation on multi-modal features to achieve fine-grained cross-modal information fusion. Our method not only retains the accuracy of Transformer in feature capture, but also effectively reduces the computational complexity through low-rank decomposition, avoiding the overfitting problem that may be caused by traditional methods.} 

{Overall, we introduce a novel and lightweight cross-modal feature fusion method called Low-Rank Matching Attention Method (LMAM). LMAM effectively captures contextual emotional semantic information in conversations while mitigating the quadratic complexity issue caused by the self-attention mechanism. Different from the existing Transformers with a self-attention mechanism, the number of parameters required by LMAM is less than one-third of the self-attention mechanism. Furthermore, LMAM is much easier to learn and may also reduce the risk of over-fitting. Fig.~\ref{figure1} illustrates a comparison of different cross-modal fusion methods. }

{Specifically, we first use the text, video and audio features pre-extracted by BERT, 3D-CNN and openSMILE respectively to obtain the corresponding query feature vector through a linear layer. In order to reduce the amount of parameters required by the network, we introduce low-rank weight decomposition to achieve compression of the learnable parameters $W$ of the linear layer. Then we perform row-by-row attention calculations on the text query feature vector, video query feature vector, and audio query feature vector with the other two modal features (i.e., matching feature vectors). Finally, we fuse the obtained multi-modal features to obtain the final high-level discourse representation with emotional contextual semantic information.}

Overall, the contributions of this work are summarized as:

\begin{itemize}
	\item We introduce a novel and lightweight cross-modal feature fusion method called Low-Rank Matching Attention Method (LMAM). LMAM effectively captures contextual emotional semantic information in conversations while mitigating the quadratic complexity issue caused by the self-attention mechanism.
	
	\item By introducing only one learnable matching feature vectors and further utilizing the low-rank decomposition method, LMAM can significantly reduce the model complexity of existing Transformer-based cross-modal fusion methods. Furthermore, LMAM can also reduce the risk of overfitting.
	
	\item Extensive experiments also verify that the proposed LMAM method can be embedded into the existing DL-based CER methods to improve their recognition accuracy in a plug-and-play manner. In addition, LMAM is a general cross-modal feature fusion method and has potential application value in other multi-modal feature fusion tasks.
\end{itemize}

\section{Related work}

\subsection{Conversational Emotion Recognition}
Conversational emotion recognition (CER) involves cross-field knowledge such as cognitive science and brain science, and has received extensive attention from researchers \cite{yang2022hybrid, liu2022social, zhang2024cross}.  Current CER research mainly includes three directions, i.e., sequential context-based emotion recognition, distinguishing speaker state-based emotion recognition, and speaker information-based emotion recognition \cite{li2022contrast}.

For the sequential context-based CER approaches, Poria et al. \cite{poria2017context} proposed a Bidirectional LSTM (bcLSTM), which utilizes recurrent neural units to extract the speaker’s context information in the video, audio, and text features, and then uses the attention mechanism to fusion the information. Hazarika et al. \cite{hazarika2018icon} designed an interactive conversational memory network (ICON) to extract the multi-modal features of different speakers following the idea of hierarchical modeling, and then input them into the global attention network for information fusion. Xing et al. \cite{9128015} introduced an Adapted Dynamic Memory Network (A-DMN) to fine-grainedly model the dependencies between contextual utterances. Shahin et al. \cite{shahin2022novel} proposed a dual-channel long short-term memory compressed-CapsNet to improve the hierarchical representation of contextual information.

For the different speaker states-based CER methods, Majumder et al.~\cite{majumder2019dialoguernn} proposed a DialogueRNN with three gating neural units (i.e., global GRU, party GRU and emotion GRU) to encode and update context and speaker information. Hu et al. \cite{hu2021dialoguecrn} proposed a Contextual Reasoning Networks (CRN) to distinguish the speaker's emotional changes in the perceptual stage and the cognitive stage.

For the speaker information-based CER methods, Ghosal et al. \cite{ghosal2019dialoguegcn}  proposed a DialogueGCN to model the dialogue relationship between speakers by constructing a speaker relationship graph from the concatenated multi-modal feature vectors. Sheng et al. \cite{sheng2020summarize} designed a summarization and aggregation graph inference network (SumAggGIN) to consider global inferences related to dialogue topics and local inferences with adjacent utterances. Hu et al. \cite{hu2021dialoguecrn} proposed a dialogue contextual reasoning network (DCRN) to extract contextual information from a cognitive perspective, and designed a multi-round reasoning module to fuse the emotional information. {The DCDM proposed by Su et al. \cite{su2024dynamic} introduced a causal directed acyclic graph (DAG) structure to establish complex correlations between hidden emotional information and other observed dialogue elements and construct a dynamic time disentanglement model to capture the hidden emotional change information in the dialogue.}

\subsection{Multimodal Feature Fusion Approaches}
In this subsetion, we briefly review the multi-modal feature fusion methods for the CER task. Liu et al. \cite{Liu2018EfficientLM} designed a low-rank multi-modal fusion method (LFM) to reduce the computational complexity caused by the change of tensor dimensions. Hu et al. \cite{hu2021mmgcn} proposed a multi-modal fused graph convolutional network (MMGCN) to model dialogue relations between speakers and fuse the cross-modal features. Lian et al. \cite{lian2021ctnet} proposed a Conversational Transformer Network to fuse complementary semantic information from different modalities. Hu et al. \cite{hu2022mm} proposed Multimodal Dynamic Fusion Network (MM-DFN), which performs emotion recognition by eliminating contextual redundant information. {Ma et al. \cite{10109845} proposed a transformer-based self-distillation model to dynamically learn the intra-modal and inter-modal information interaction.} {Li et al. \cite{li2023graphmft} proposed a Graph network based Multi-modal Fusion Technology (GraphMFT) for CER. GraphMFT achieves the fusion of intra-modal and inter-modal contextual semantic information by building multiple graph attention networks.} {The CFN-ESA proposed by Li et al. \cite{li2024cfn} effectively integrated the data distribution between modalities and promotes the interaction of multimodal information by constructing a unimodal encoder, a cross-modal encoder and a sentiment conversion module.} Although these multi-modal fusion approaches can obtain discriminative fused feature by exploiting the information of different modalities, they are either computationally expensive or do not fully consider the complementary information of different modals.

\section{Preliminary Information}

\subsection{Problem Definition}
We assume the participants in the dialogue are $P=\{p_1,p_2,\ldots,p_N \}$, where $N$ represents the number of participants $(N\geq2)$. We define sequential context $T=\{t_1,t_2,\ldots,t_M \}$, where $M$ represents the total number of sessions and $t_i$ represents the $i$-th utterance. The task of CER is to identify the discrete emotions (e.g., happy, sad, disgust, neutral, excited, etc.) in each utterance.

\subsection{Multimodal Feature Extraction}
In the CER task, three types of modality data are included, namely text, video and audio. The feature extraction method of each modal is different, and the semantic information they contain is also different \cite{ghosh2022comma}. Next, we will briefly introduce their data preprocessing methods.

\textbf{Word Embedding:} To obtain the feature vector representation of characters that computers can understand \cite{schuster2022bert}, we use the large-scale pre-training model BERT to encode text features. First, we use the Tokenizer method to segment the text to get each word and its index. We then feed them into the BERT model for feature encoding, and use the first 100-dimensional features in the BERT model as our text feature vectors.

\textbf{Visual Feature Extraction:} Following Hazarika et al. \cite{hazarika2018conversational}, we use 3D-CNN to extract the speaker's facial expression features and gesture change features in video frames, which have an important impact on the model's understanding of the speaker's emotional changes. Specifically, we utilize 3D-CNN and a linear layer with 512 neurons to obtain video feature vectors with rich semantic information.

\textbf{Audio Feature Extraction:} Following Hazarika et al. \cite{hazarika2018conversational}, we use openSMILE \cite{eyben2010opensmile} to extract acoustic features in audio (e.g., loudness, Mel-spectra, MFCC). Specifically, we utilize the $IS12\_ComParE1$ extractor\footnote{ http://audeering.com/technology/opensmile}  in openSMILE and a linear layer with 100 neurons to obtain speaker audio features.

{\subsection{Self-attention Mechanism}}
{Self-Attention Mechanism is an algorithm used for sequence data processing to calculate the correlation between various positions of the input sequence. Owing to its powerful information capture capabilities, self-attention mechanisms have received considerable research attention in various fields, e.g., natural language understanding and text generation. A typical attention mechanism is to obtain the attention score by calculating the correlation between each element in the input sequence and other elements. For a given input sequence $\boldsymbol{X}$, query $\boldsymbol{Q}$, key $\boldsymbol{K}$ and value $\boldsymbol{V}$ are calculated as follows:}
\begin{equation}
    \boldsymbol{Q},\boldsymbol{K},\boldsymbol{V} = \boldsymbol{W}^Q\boldsymbol{X},\boldsymbol{W}^K\boldsymbol{X}, \boldsymbol{W}^V\boldsymbol{X}
\end{equation}
{where $\boldsymbol{Q}$, $\boldsymbol{K}$,and $\boldsymbol{V}$ are the learnable parameters.}

{Given the query $\boldsymbol{Q}$, key $\boldsymbol{K}$ and value $\boldsymbol{V}$, the embeddings representation of the sequence are obtained through the dot product and Softmax function as follows:}
\begin{equation}
\text{Attention}
(\boldsymbol{Q},\boldsymbol{K},\boldsymbol{V})=\text{softmax}\left(\frac{\boldsymbol{QK}^\top}{\sqrt{d_k}}\right)\boldsymbol{V}
\end{equation}
{In this work, we only use one query $\boldsymbol{Q}$ as our matching feature vectors for self-attention calculation since the high complexity of the model easily leads to the risk of overfitting.}

\section{Proposed LMAM Cross-modal Fusion Method}
In this section, we propose a novel cross-modal fusion method, namely Low-rank Matching Attention Mechanism (LMAM).

\subsection{Matching Attention Layer}
We concatenate the row vectors of all 1s with the extracted multi-modal features to obtain $\xi$ as follows:
\begin{equation}
	\boldsymbol{\xi}=\left\{\boldsymbol{\xi}_u, \boldsymbol{\xi}_a, \boldsymbol{\xi}_v\right\},
\end{equation}
where $\boldsymbol{\xi}_u$ represents context utterence features, $\boldsymbol{\xi}_a$ represents audio features, and $\boldsymbol{\xi}_v$ represents video features.

The existing CER approaches usually use feature splicing or feature summation to fuse the cross-modal feature \cite{zadeh2017tensor,Liu2018EfficientLM,hu2021mmgcn,Liu2018EfficientLM}. As introduced in the related work, these cross-modal fusion methods are either computationally expensive or do not fully consider the complementary information of different modals. Therefore, our goal is to construct an efficient and effective fusion method that captures the differences among multimodal features by computing the correlation among the three modalities of text, video and audio and realizes the fusion of complementary semantic information across modalities. Specifically, the computation process of our proposed LMAM fusion method is shown as follows.

\begin{figure*}
	\centering
	\includegraphics[width=0.98\linewidth]{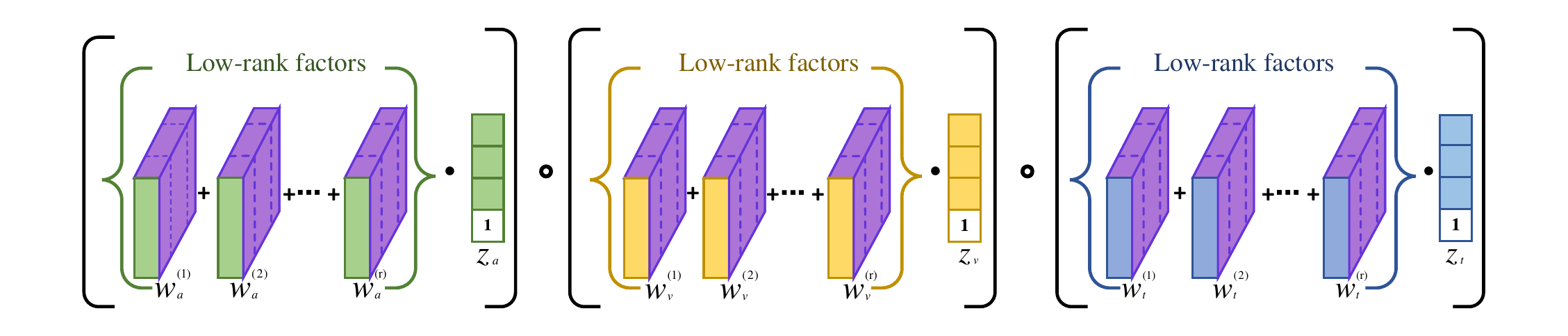}
	\caption{LMAM achieves information fusion of multimodal features through parallel low-rank decomposition of weight (i.e., $\omega_a, \omega_v, \omega_t$) and modal features. We add an extra dimension to each modality feature and pad them with 1 to ensure that the intra-modal semantic information is preserved during inter-modal feature fusion.}
	\label{fig:low-rank-attention}
\end{figure*}

\begin{algorithm}[tb]
	\caption{Matching Attention Mechanism (LMAM)}
	\label{alg:algorithm}
	\textbf{Input}: Text feature vectors $\xi_t$, video feature vectors $\xi_v$ and audio feature vectors $\xi_a$; the number of iterations $\epsilon$; the size of dataset $\phi$.\\ 
	
	\textbf{Output}: The enhanced multimodal fusion feature vectors $\xi_f$.
	
	\begin{algorithmic}[1] 
		\STATE Initialize the model weights $W$ and bias $b$.
		\STATE Initialize the set of multimodal feature fusion $\xi_{fusion}$.
		\FOR{EPOCH $\leftarrow 1,2,\ldots,\epsilon$}
		\FOR{$i \leftarrow 1,2,\ldots,len(\phi) / 32$ }
		\STATE Sample a batch $\xi=\{\xi_t, \xi_v, \xi_a\}_{i=1}^{32}$.
		\FOR{modal in $\xi$}
		\FOR{t in modal}
		\STATE att\_emo, score=MatchingAttention(modal, t)
		\STATE att\_emotions.append(att\_em.unsqueeze(0))
		\STATE scores.append(score[:, 0, :])
		\ENDFOR
		\STATE att\_emotions=torch.cat(att\_emotions, dim=0)
		\STATE $\varPsi$   =att\_emotions + F.gelu(modal)
		\ENDFOR
		\ENDFOR
		\ENDFOR
		\STATE $\xi_{fusion}=\{\varPsi_1,\varPsi_2,\ldots,\varPsi_{len(\phi)/32}\}$.
		\STATE \textbf{return} the enhanced multimodal feature vectors $\xi_{fusion}$.
	\end{algorithmic}
\end{algorithm}

{For a given model input $\boldsymbol{K}_i$, $\boldsymbol{V}_i$ and $\boldsymbol{M}_i$, {where $\boldsymbol{K}_i = \boldsymbol{\xi}_i$, $\boldsymbol{V}_i = \boldsymbol{\xi}_i$, and $\boldsymbol{M}_i = \boldsymbol{\xi}_i$.} We first compute the query matrix $\boldsymbol{Q}_i \in \mathbb{R}^{L_{\boldsymbol{Q}_i} \times d_{\boldsymbol{Q}_i}}$ {by linear transformation from $\boldsymbol{M}_i$} as follows:}
\begin{equation}
	\boldsymbol{Q}_i = \boldsymbol{M}_i\boldsymbol{W}^{{Q}_i}+\boldsymbol{b}^{{Q}_i},
\end{equation}
{where $\boldsymbol{K}_i$ and $\boldsymbol{M}_i$ represent the $i$-th modal features. {$L_{{Q}_i}$} represents the sequence length of the modal features. $d_{{Q}_i}$ represents the feature dimension after linear layer mapping, and $\boldsymbol{W}^{{Q}_i} \in \mathbb{R}^{d_m \times d_{{Q}_i}}$, $d_m$ represents the feature embedding dimension of $\boldsymbol{Q}_i$.}
 
%

Next, we get the attention score using the following formula:
\begin{equation}	
	\begin{aligned}
	\boldsymbol{\alpha}_i=softmax\left(Tanh\left(\frac{\boldsymbol{Q}_i \boldsymbol{K}_i^\top}{\sqrt{d_k}}\right)\right) \\
	\end{aligned}
\end{equation}
{where $\top$ represents the matrix transpose operation. $d_k$  represents the sequence length of the modal features, $\boldsymbol{\alpha}_i$ represents the attention score, and $\boldsymbol{\alpha}_i \in [0,1]$.}

Subsequently, we perform matrix multiplication by the attention score and the modality feature $\boldsymbol{I}_i$ to obtain the attention output as follows:
\begin{equation}
	\begin{aligned}
	\boldsymbol{A}_i =\boldsymbol{\alpha_i\boldsymbol{V}_i}
	\end{aligned}
\end{equation}
{where {$\boldsymbol{A}_i$} is the feature vector after attention calculation. As shown in Fig. \ref{fig:transformer}, we present the fusion process of text features and audio features using a matched attention mechanism. Similarly, the fusion process for text and video, and video and audio follows the same paradigm.}

\begin{figure}
	\centering
	\includegraphics[width=0.9\linewidth]{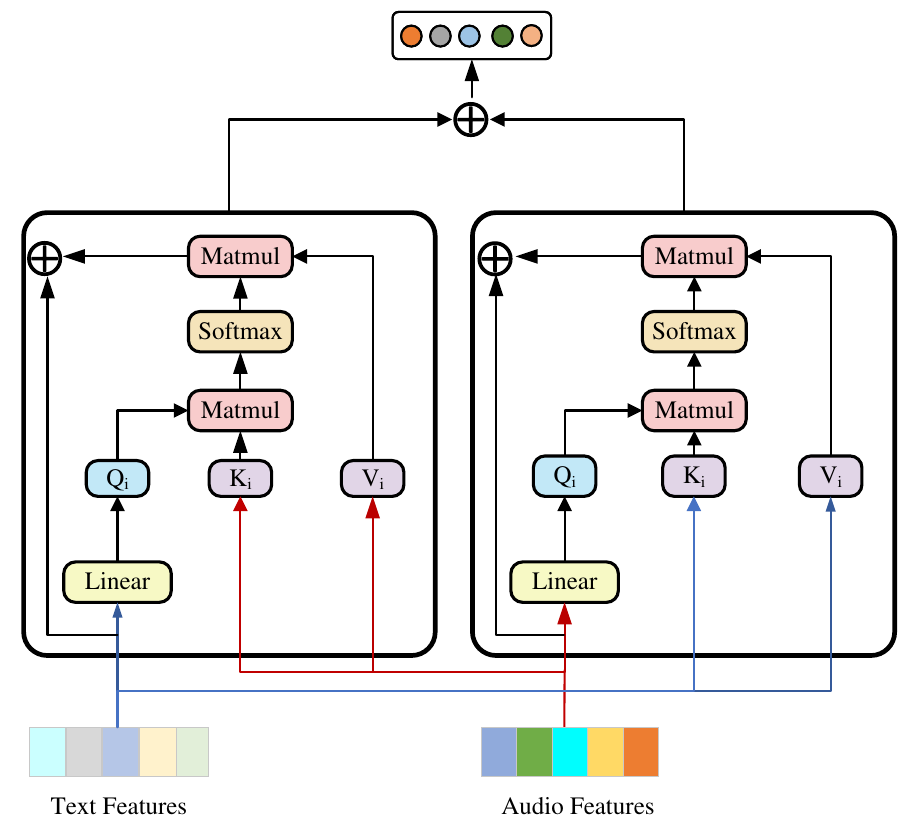}
	\caption{An illustrative example of the proposed LMAM method fusing complementary semantic information between different modal features. We show the fusion process of text features and audio features. The fusion process for text and video, and video and audio follows the same paradigm.}
	\label{fig:transformer}
\end{figure}

{To prevent the problem of gradient disappearance and information collapse in the model training, we also build a residual connection layer with normalization operation. Finally, we use a linear layer with ReLU activation function to get the final output of the LMAM. The formulas are as follows:}
\begin{equation}
\begin{aligned}
	&\boldsymbol{E}_i = Norm(\boldsymbol{A}_i + \boldsymbol{I}_i), \\
	&\boldsymbol{O}_i = ReLU(Linear(\boldsymbol{E}_i)).
 \end{aligned}
\end{equation}

\subsection{Low-rank Weight Decomposition}

Low-rank weight decomposition technology achieves parameter reduction and model compression by decomposing the original weight matrix into the product of two or more low-rank matrices \cite{kolda2009tensor}. Since it can effectively reduce the complexity of the model, low-rank decomposition technology has received extensive research attention on many tasks, e.g., image processing and multi-modal fusion. {Tucker decomposition is a commonly used technique for tensor decomposition, which can decompose a high-order tensor into a smaller core tensor and a set of factor matrices. Unlike other tensor decomposition methods, Tucker decomposition allows the core tensor to maintain complex relationships between multiple modes, thus being able to capture multimodal dependencies in the data.}

\begin{figure*}
	\centering
	\includegraphics[width=1\linewidth]{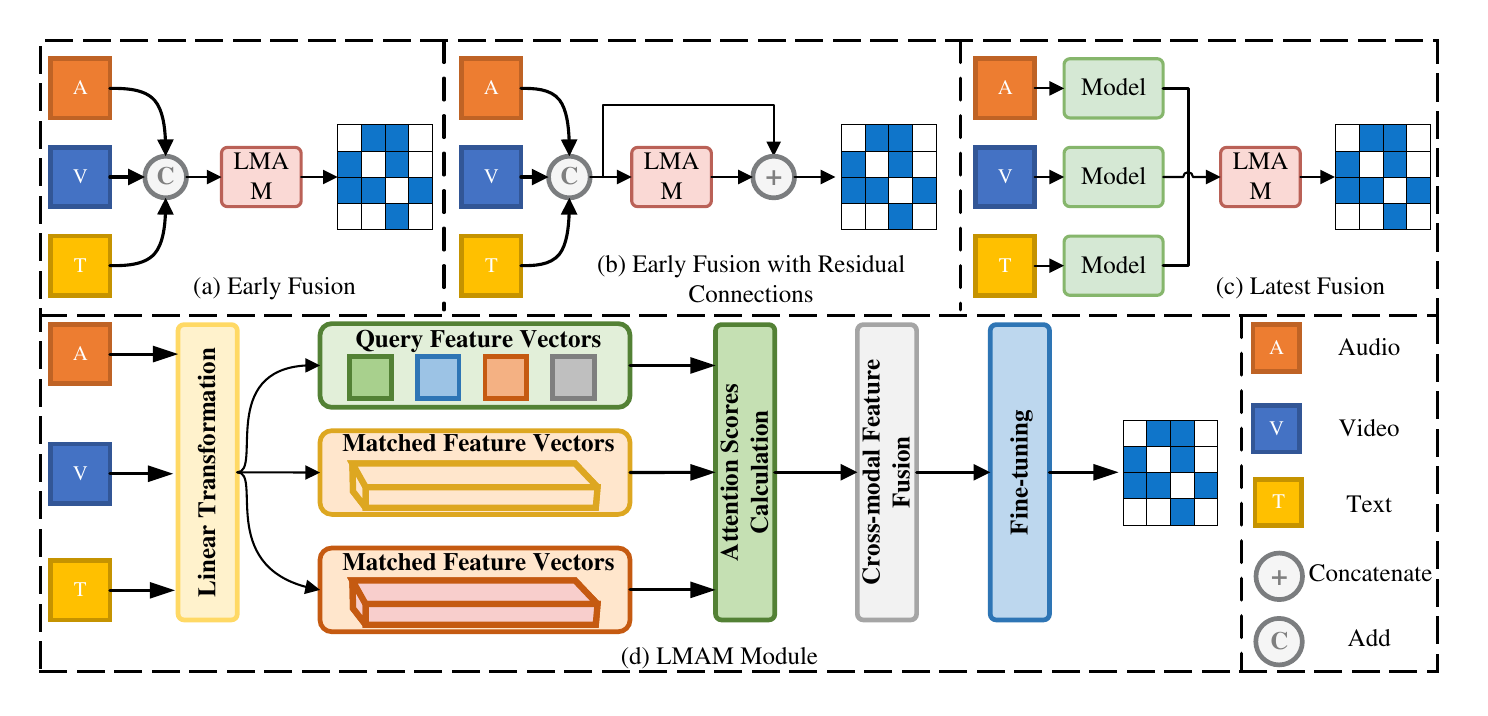}
	\caption{Three embedding ways for cross-modal fusion using LMAM module. (a) Embedding the LMAM module before model's feature extraction. (b) Embedding the LMAM module and using the residual connections before model's feature extraction. (c) Embedding the LMAM module after model's feature extraction. (d) The overall flow of the LMAM module. {$\boldsymbol{Q}_i$ is represented as the query feature vectors, and $\boldsymbol{I}_i$ is represented as the matching feature vectors.}}
	\label{fig:attention}
\end{figure*}

{As shown in Fig. \ref{fig:low-rank-attention}, the idea behind low-rank decomposition in LMAM is to decompose the {weight $\boldsymbol{W}$} into specific factors that match the modal features. For any $N$-order weight $\boldsymbol{W}_i$, there is always a low-rank decomposition method. The formula is defined as follows:}
\begin{equation}
	\begin{aligned}
	\widetilde{\boldsymbol{W}}_i&=Tucker(\boldsymbol{W}_i) \\
	&=\sum_{j=1}^R \omega_{n, i}^{(j)}, \omega_{n, i}^{(j)} \in \mathbb{R}_n^d
	\end{aligned}
\end{equation}
{where $r$ represents the rank of the weight $W_i$, {$\left\{\left\{\omega_{n, i}^{(j)}\right\}_{n=1}^\eta\right\}_{j=1}^r$} is a collection of low-rank decomposition factors, and $\eta$ is the size of the multimodal feature set $\xi=\{\xi_u,\xi_a,\xi_v\}$, i.e., $\eta=3$. $\widetilde{\boldsymbol{W}}_i$ is computed during training.}

{Therefore, Eq. 4 can be calculated as:}
\begin{equation}
	\boldsymbol{Q}_i=\left(\sum_{j=1}^r \omega_{n, i}^{(j)}\right)\boldsymbol{M}_i+\boldsymbol{b}^{\boldsymbol{Q}_i}, \omega_{n, i}^{(j)} \in \mathbb{R}_n^d
\end{equation}

The whole computational process of the LMAM method is shown in Figure~\ref{figure1}(c) and the pseudocode of the LMAM method is summarized in Algorithm 1.

\subsection{Comparison to Self-attention}
Studies \cite{dosovitskiyimage} have shown that the performance of the self-attention mechanism is lower than CNN, RNN and other methods when the amount of data is small, while its performance can gradually exceed CNN and RNN when the amount of data is very large. The difference in performance may be attributed to  that the self-attention mechanism needs to learn the query vectors $\boldsymbol{Q}$, the key vectors $\boldsymbol{K}$, and the value vectors $\boldsymbol{V}$ at the same time, which makes the optimization of the model more difficult. Unlike classic self-attention, LMAM only needs a very low rank weight to achieve better performance than self-attention. Specifically, we only set a learnable parameter $\boldsymbol{W^Q}$ for cross-modal feature fusion and capture of complementary semantic information. Furthermore, we perform a parallel low-rank decomposition of $\boldsymbol{W^Q}$ with modality-specific factors to further reduce the amount of parameters required for $\boldsymbol{W^Q}$. LMAM can reduce the difficulty of network optimization while maintaining performance.

\begin{table}[htbp]
	{\caption{The division of training set, validation set and test set of IEMOCAP and MELD data sets.}}
	\renewcommand\arraystretch{1.2}
	\setlength{\tabcolsep}{3mm}{
		{
			\begin{tabular}{cccc}
				\toprule
				Datasets                 & Partition & Utterance Count & Dialogue Count \\ \midrule
				\multirow{2}{*}{IEMOCAP} & train+val & 5810            & 120            \\
				& test      & 1623            & 31             \\
				\multirow{2}{*}{MELD}    & train+val & 11098           & 1153           \\
				& test      & 2610            & 280  \\
				\multirow{2}{*}{CMU-MOSI}    & train+val & 1513           &    62        \\
				& test      & 686            & 31 \\
				\multirow{2}{*}{POM}    & train+val & 700           &    N/A        \\
				& test      & 203            & N/A
				\\ \bottomrule
	\end{tabular}}}
\end{table}

\begin{table*}[!t]
	
	\renewcommand\arraystretch{1.1}
	\caption{Experimental results on IEMOCAP dataset. Methods with $\ast$ represent the method equipped with our LMAM module without any further changes. The best result in each column is in bold.}
	\label{table1}
	\renewcommand\arraystretch{1.2}
	\setlength{\tabcolsep}{3mm}{
		\begin{tabular}{l|ccccccc}
			\hline
			\multirow{3}{*}{Methods} & \multicolumn{7}{c}{IEMOCAP}                                                              \\ \cline{2-8}
			& Happy      & Sad        & Neutral    & Angry      & Excited    & Frustrated & Average(w) \\ \cline{2-8}
			& Acc.  F1   & Acc.  F1   & Acc.  F1   & Acc.  F1   & Acc.  F1   & Acc.  F1   & Acc.  F1   \\ \hline
			TextCNN \cite{2014Convolutional}                      & 27.73  29.81 & 57.14  53.83 & 34.36  40.13 & \textbf{61.12}  52.47 & 46.11  50.09 & \textbf{62.94}  55.78 & 48.89  48.08 \\
			\rowcolor{gray!30}
			TextCNN$^\ast$ & \textbf{37.24 37.37} & \textbf{81.41 72.97} & \textbf{50.57 53.95} & 57.21 \textbf{62.96} & \textbf{70.28 63.87} & 59.01 \textbf{59.16} & \textbf{60.35 59.34}\\ 
			$\Delta$ &  \textcolor[rgb]{0.0,0.6,0.0}{$\uparrow$9.51} \textcolor[rgb]{0.0,0.6,0.0}{$\uparrow$7.56}  &  \textcolor[rgb]{0.0,0.6,0.0}{$\uparrow$24.27} \textcolor[rgb]{0.0,0.6,0.0}{$\uparrow$19.14}  &\textcolor[rgb]{0.0,0.6,0.0}{$\uparrow$16.21} \textcolor[rgb]{0.0,0.6,0.0}{$\uparrow$13.82}   &\textcolor[rgb]{0.2,0.2,0.2}{$\downarrow$3.91} \textcolor[rgb]{0.0,0.6,0.0}{$\uparrow$10.49}   &\textcolor[rgb]{0.0,0.6,0.0}{$\uparrow$24.17} \textcolor[rgb]{0.0,0.6,0.0}{$\uparrow$13.78}  &
			\textcolor[rgb]{0.2,0.2,0.2}{$\downarrow$3.87} \textcolor[rgb]{0.0,0.6,0.0}{$\uparrow$3.29}  &\textcolor[rgb]{0.0,0.6,0.0}{$\uparrow$11.46} \textcolor[rgb]{0.0,0.6,0.0}{$\uparrow$11.26}   \\ \hline

			bc-LSTM \cite{poria2017context}                 & 29.16  34.49 & 57.14  60.81 & 54.19  51.80 & 57.03  56.75 & 51.17  57.98 &\textbf{ 67.12}  58.97 & 55.19  54.96 \\
			bc-LSTM+Att \cite{poria2017context} & 30.56 \textbf{35.63} & 56.73 62.09 & 57.55 53.00 & {59.41} \textbf{59.24} & 52.84 58.85 & 65.88 \textbf{59.41} & 56.31 56.08  \\
			\rowcolor{gray!30}
			bc-LSTM$^\ast$ & \textbf{86.67} 31.52 & \textbf{66.43 73.08} & \textbf{63.09 55.54} & \textbf{81.51} 49.79 & \textbf{68.08 77.01} & 50.39 59.31 & \textbf{65.55 60.29}\\ 
			$\Delta$ &  \textcolor[rgb]{0.0,0.6,0.0}{$\uparrow$56.11} \textcolor[rgb]{0.2,0.2,0.2}{$\downarrow$4.11}  & \textcolor[rgb]{0.0,0.6,0.0}{$\uparrow$9.7} \textcolor[rgb]{0.0,0.6,0.0}{$\uparrow$10.99}   & \textcolor[rgb]{0.0,0.6,0.0}{$\uparrow$5.54} \textcolor[rgb]{0.0,0.6,0.0}{$\uparrow$2.54}   & \textcolor[rgb]{0.0,0.6,0.0}{$\uparrow$22.1} \textcolor[rgb]{0.2,0.2,0.2}{$\downarrow$9.45}   & \textcolor[rgb]{0.0,0.6,0.0}{$\uparrow$15.24} \textcolor[rgb]{0.0,0.6,0.0}{$\uparrow$18.16}  & \textcolor[rgb]{0.2,0.2,0.2}{$\downarrow$15.49} \textcolor[rgb]{0.2,0.2,0.2}{$\downarrow$0.1}  & \textcolor[rgb]{0.0,0.6,0.0}{$\uparrow$10.36} \textcolor[rgb]{0.0,0.6,0.0}{$\uparrow$5.33} \\ \hline

			DialogueRNN   \cite{majumder2019dialoguernn}           & 25.63  33.11 & \textbf{75.14  78.85} & 58.56  {59.24} & 64.76 \textbf{65.23} & \textbf{80.27}  71.85 & \textbf{61.16}  58.97 & 63.40  62.77 \\
			\rowcolor{gray!30}
			DialogueRNN$^\ast$ & \textbf{63.51 66.43} & 72.43 78.64 & \textbf{62.94 59.53} & \textbf{78.89} 55.04 & 77.00 \textbf{77.65} & 55.07 \textbf{59.07} & \textbf{66.84 65.79}\\ 
			$\Delta$ & \textcolor[rgb]{0.0,0.6,0.0}{$\uparrow$37.88} \textcolor[rgb]{0.0,0.6,0.0}{$\uparrow$33.32}   & \textcolor[rgb]{0.2,0.2,0.2}{$\downarrow$2.71} \textcolor[rgb]{0.2,0.2,0.2}{$\downarrow$0.21}   & \textcolor[rgb]{0.2,0.2,0.2}{$\uparrow$4.38} \textcolor[rgb]{0.0,0.6,0.0}{$\uparrow$0.29}   &  \textcolor[rgb]{0.0,0.6,0.0}{$\uparrow$14.13} \textcolor[rgb]{0.2,0.2,0.2}{$\downarrow$10.19}  & \textcolor[rgb]{0.2,0.2,0.2}{$\downarrow$3.27} \textcolor[rgb]{0.2,0.2,0.2}{$\uparrow$5.8}  & \textcolor[rgb]{0.2,0.2,0.2}{$\downarrow$6.09} \textcolor[rgb]{0.2,0.2,0.2}{$\uparrow$0.1}  & \textcolor[rgb]{0.0,0.6,0.0}{$\uparrow$3.44} \textcolor[rgb]{0.0,0.6,0.0}{$\uparrow$3.02}  \\  \hline

			DialogueGCN   \cite{ghosal2019dialoguegcn}           & 40.63  42.71 & \textbf{89.14  84.45} & 61.97  \textbf{63.54} & 67.51  \textbf{64.14} & 65.46  63.08 & \textbf{64.13  66.90} & 65.91  65.62 \\
			\rowcolor{gray!30}
			DialogueGCN$^\ast$ & \textbf{62.60 59.69} & 76.86 78.98 & \textbf{62.82} 60.06 & \textbf{74.40} 63.48 & \textbf{72.38 77.93} & 56.40 58.30 & \textbf{66.39 66.12}\\ 
			$\Delta$ & \textcolor[rgb]{0.0,0.6,0.0}{$\uparrow$21.97} \textcolor[rgb]{0.0,0.6,0.0}{$\uparrow$16.98}   & \textcolor[rgb]{0.2,0.2,0.2}{$\downarrow$12.28} \textcolor[rgb]{0.2,0.2,0.2}{$\downarrow$5.47}   &  \textcolor[rgb]{0.0,0.6,0.0}{$\uparrow$0.85} \textcolor[rgb]{0.2,0.2,0.2}{$\downarrow$3.48}  &  \textcolor[rgb]{0.0,0.6,0.0}{$\uparrow$6.89} \textcolor[rgb]{0.2,0.2,0.2}{$\downarrow$0.66}  & \textcolor[rgb]{0.0,0.6,0.0}{$\uparrow$6.92} \textcolor[rgb]{0.0,0.6,0.0}{$\uparrow$14.85}  & \textcolor[rgb]{0.2,0.2,0.2}{$\downarrow$7.73} \textcolor[rgb]{0.2,0.2,0.2}{$\downarrow$8.6}  & \textcolor[rgb]{0.0,0.6,0.0}{$\uparrow$0.48} \textcolor[rgb]{0.0,0.6,0.0}{$\uparrow$0.51}  \\  \hline
			
			MM-DFN \cite{hu2022mm}& {40.17 42.22} & \textbf{74.27} 78.98 & \textbf{69.13 66.42} & {70.25} \textbf{69.97} & 76.99 75.56 & \textbf{68.58 66.33} & {68.77 68.20}\\
			\rowcolor{gray!30}
			MM-DFN$^\ast$ & \textbf{74.58 69.57} & 74.01 \textbf{81.03} & 64.93 63.71 & \textbf{73.91} 66.67 & \textbf{81.40 80.00} & 59.84 61.71 & \textbf{69.94 69.69}\\ 
			$\Delta$ &  \textcolor[rgb]{0.0,0.6,0.0}{$\uparrow$34.41} \textcolor[rgb]{0.0,0.6,0.0}{$\uparrow$27.35}  &  \textcolor[rgb]{0.2,0.2,0.2}{$\downarrow$0.26} \textcolor[rgb]{0.0,0.6,0.0}{$\uparrow$2.05}  &  \textcolor[rgb]{0.2,0.2,0.2}{$\downarrow$4.2} \textcolor[rgb]{0.2,0.2,0.2}{$\downarrow$2.71}  &  \textcolor[rgb]{0.0,0.6,0.0}{$\uparrow$3.66} \textcolor[rgb]{0.2,0.2,0.2}{$\downarrow$3.3}  & \textcolor[rgb]{0.0,0.6,0.0}{$\uparrow$4.41} \textcolor[rgb]{0.0,0.6,0.0}{$\uparrow$4.44}  & \textcolor[rgb]{0.2,0.2,0.2}{$\downarrow$8.74} \textcolor[rgb]{0.2,0.2,0.2}{$\downarrow$4.62}  & \textcolor[rgb]{0.0,0.6,0.0}{$\uparrow$1.16} \textcolor[rgb]{0.0,0.6,0.0}{$\uparrow$1.48}  \\  \hline
			
			M2FNet \cite{chudasama2022m2fnet}& {65.92 60.00} & \textbf{79.18 82.11} & 65.80 \textbf{65.88} & \textbf{75.37 68.21} & {74.84 72.60} & \textbf{59.87 68.31} & {69.11 69.86}\\
			\rowcolor{gray!30}
			M2FNet$^\ast$ & \textbf{73.35 69.53} & 74.37 81.42 & \textbf{67.26} 63.51 & {72.86} 66.23 & \textbf{81.57 81.29} & 59.64 62.50 & \textbf{70.31 70.07}\\ 
			$\Delta$ & \textcolor[rgb]{0.0,0.6,0.0}{$\uparrow$7.43} \textcolor[rgb]{0.0,0.6,0.0}{$\uparrow$9.53}  &  \textcolor[rgb]{0.2,0.2,0.2}{$\downarrow$4.81} \textcolor[rgb]{0.2,0.2,0.2}{$\downarrow$0.69}  & \textcolor[rgb]{0.0,0.6,0.0}{$\uparrow$1.46} \textcolor[rgb]{0.2,0.2,0.2}{$\downarrow$2.37}   & \textcolor[rgb]{0.2,0.2,0.2}{$\downarrow$2.51} \textcolor[rgb]{0.2,0.2,0.2}{$\downarrow$1.98}   & \textcolor[rgb]{0.0,0.6,0.0}{$\uparrow$6.73} \textcolor[rgb]{0.0,0.6,0.0}{$\uparrow$8.69}  & \textcolor[rgb]{0.2,0.2,0.2}{$\downarrow$0.23} \textcolor[rgb]{0.2,0.2,0.2}{$\downarrow$5.81}  & \textcolor[rgb]{0.0,0.6,0.0}{$\uparrow$1.20} \textcolor[rgb]{0.0,0.6,0.0}{$\uparrow$0.21} \\   \hline
			
			EmoCaps \cite{li2022emocaps}& {\textbf{70.34}} {72.86} & 77.39 82.45 & 64.27 65.10 & 71.79 \textbf{69.14} & 84.50 73.90 &\textbf{63.94} 63.41 & {71.22 70.06} \\
			\rowcolor{gray!30}
			EmoCaps$^\ast$ & {69.93} {\textbf{74.31}} & \textbf{82.52 85.47} & \textbf{68.41 67.03} & \textbf{79.49} 65.26 & \textbf{84.85 80.14} & 63.33 \textbf{68.38} & \textbf{73.67 73.01}\\ 
			$\Delta$ &  \textcolor[rgb]{0.2,0.2,0.2}{$\downarrow$0.41} \textcolor[rgb]{0.0,0.6,0.0}{$\uparrow$1.45}  &  \textcolor[rgb]{0.0,0.6,0.0}{$\uparrow$5.13} \textcolor[rgb]{0.0,0.6,0.0}{$\uparrow$3.02}  &  \textcolor[rgb]{0.0,0.6,0.0}{$\uparrow$4.14} \textcolor[rgb]{0.0,0.6,0.0}{$\uparrow$1.93}  & \textcolor[rgb]{0.0,0.6,0.0}{$\uparrow$7.7} \textcolor[rgb]{0.2,0.2,0.2}{$\downarrow$3.88}   & \textcolor[rgb]{0.0,0.6,0.0}{$\uparrow$0.35} \textcolor[rgb]{0.0,0.6,0.0}{$\uparrow$6.24}  & \textcolor[rgb]{0.2,0.2,0.2}{$\downarrow$0.61} \textcolor[rgb]{0.0,0.6,0.0}{$\uparrow$4.97}  & \textcolor[rgb]{0.0,0.6,0.0}{$\uparrow$2.45} \textcolor[rgb]{0.0,0.6,0.0}{$\uparrow$2.95} \\   \hline
	\end{tabular}}
\end{table*}

\begin{table*}[!t]
	
	\caption{Experimental results on MELD dataset. Methods with $\ast$ represent the method equipped with our LMAM module without any further changes. The best result in each column is in bold.}
	\label{table2}
	\renewcommand\arraystretch{1.2}
	\setlength{\tabcolsep}{7.7pt}{
		\begin{tabular}{l|cccccccc}
			\hline
			\multirow{3}{*}{Methods} & \multicolumn{8}{c}{MELD}                                                                            \\ \cline{2-9}
			& Neutral     & Surprise    & Fear     & Sadness    & Joy        & Disgust  & Anger      & Average(w) \\ \cline{2-9}
			& Acc.  F1    & Acc.  F1    & Acc.  F1 & Acc.  F1   & Acc.  F1   & Acc.  F1 & Acc.  F1   & Acc. F1   \\ \hline
			TextCNN \cite{2014Convolutional}                      & \textbf{71.23}  74.91  & \textbf{43.35  45.51}  & \textbf{4.63  3.71} & 18.25  21.17 & 46.14  49.47 & \textbf{8.91  8.36} & 35.33  34.51 & 52.48 55.09 \\
			\rowcolor{gray!30}
			TextCNN$^\ast$ & 70.23 {\textbf{75.79}} & {36.47 44.78} & 0.00 0.00 & \textbf{24.19 21.19} & \textbf{50.58 52.46} & 0.00 0.00 & \textbf{43.43 41.68} & \textbf{53.17 56.55} \\ 
			
			$\Delta$ & \textcolor[rgb]{0.2,0.2,0.2}{$\downarrow$1.00} \textcolor[rgb]{0.0,0.6,0.0}{$\uparrow$0.88} & \textcolor[rgb]{0.2,0.2,0.2}{$\downarrow$6.88} \textcolor[rgb]{0.2,0.2,0.2}{$\downarrow$0.73} & \textcolor[rgb]{0.2,0.2,0.2}{$\downarrow$4.63} \textcolor[rgb]{0.2,0.2,0.2}{$\downarrow$3.71} & \textcolor[rgb]{0.0,0.6,0.0}{$\uparrow$5.94} \textcolor[rgb]{0.0,0.6,0.0}{$\uparrow$0.02} & \textcolor[rgb]{0.0,0.6,0.0}{$\uparrow$4.44} \textcolor[rgb]{0.0,0.6,0.0}{$\uparrow$2.99} & \textcolor[rgb]{0.2,0.2,0.2}{$\downarrow$8.91} \textcolor[rgb]{0.2,0.2,0.2}{$\downarrow$8.36}  &  \textcolor[rgb]{0.0,0.6,0.0}{$\uparrow$8.10} \textcolor[rgb]{0.0,0.6,0.0}{$\uparrow$7.17} & \textcolor[rgb]{0.0,0.6,0.0}{$\uparrow$0.69} \textcolor[rgb]{0.0,0.6,0.0}{$\uparrow$1.47} \\ \hline
			
			bc-LSTM    \cite{poria2017context}              & \textbf{71.45}   73.84 & 46.82   \textbf{47.71} & \textbf{3.84  5.46} & 22.47  \textbf{25.19} & 51.61  {51.34} & \textbf{4.31  5.23} & 36.71  38.44 & 54.18  55.89 \\
			bc-LSTM+Att & 70.45 {\textbf{75.55}} & 46.43 46.35 & 0.00 0.00 &21.77 16.27 & 49.30 50.72 & 0.00 0.00 & 41.77 40.71 & 53.73 55.82 \\
			\rowcolor{gray!30}
			bc-LSTM$^\ast$ & 70.78 {75.46} & \textbf{47.18} 46.47 & 0.00 0.00 & \textbf{26.09} 24.58 & \textbf{52.33 53.11} & 0.00 0.00 & \textbf{43.23 40.92} & \textbf{54.97 56.85}\\ 
			$\Delta$ & \textcolor[rgb]{0.0,0.6,0.0}{$\uparrow$0.33} \textcolor[rgb]{0.2,0.2,0.2}{$\downarrow$0.09} & \textcolor[rgb]{0.0,0.6,0.0}{$\uparrow$0.75} \textcolor[rgb]{0.0,0.6,0.0}{$\uparrow$0.12} & \textcolor[rgb]{0.0,0.6,0.0}{$\uparrow$0.00} \textcolor[rgb]{0.0,0.6,0.0}{$\uparrow$0.00} &  \textcolor[rgb]{0.0,0.6,0.0}{$\uparrow$4.32} \textcolor[rgb]{0.0,0.6,0.0}{$\uparrow$8.31} & \textcolor[rgb]{0.0,0.6,0.0}{$\uparrow$3.03} \textcolor[rgb]{0.0,0.6,0.0}{$\uparrow$2.39} & \textcolor[rgb]{0.0,0.6,0.0}{$\uparrow$0.00} \textcolor[rgb]{0.0,0.6,0.0}{$\uparrow$0.00} & \textcolor[rgb]{0.0,0.6,0.0}{$\uparrow$1.46} \textcolor[rgb]{0.0,0.6,0.0}{$\uparrow$0.21} & \textcolor[rgb]{0.0,0.6,0.0}{$\uparrow$1.24} \textcolor[rgb]{0.0,0.6,0.0}{$\uparrow$1.03}  \\ \hline
			
			DialogueRNN  \cite{majumder2019dialoguernn}            & \textbf{72.12}   73.54 & \textbf{46.42  49.47}  & {1.61  1.23} & 23.97  \textbf{23.83} & \textbf{52.01}  50.74 & {1.52  1.73} & 41.01  {41.54} & 55.10  55.97 \\
			\rowcolor{gray!30}
			DialogueRNN$^\ast$ & 71.74 \textbf{75.76} & 45.83 48.23 & \textbf{3.13 2.77} & \textbf{31.71} 17.93 & 49.25 \textbf{53.04} & \textbf{2.01 2.58} & \textbf{42.40} {\textbf{42.21}} & \textbf{55.27 56.93}\\ 
			$\Delta$ & \textcolor[rgb]{0.2,0.2,0.2}{$\downarrow$0.38} \textcolor[rgb]{0.0,0.6,0.0}{$\uparrow$2.22} &  \textcolor[rgb]{0.2,0.2,0.2}{$\downarrow$0.59} \textcolor[rgb]{0.2,0.2,0.2}{$\downarrow$1.24} & \textcolor[rgb]{0.0,0.6,0.0}{$\uparrow$1.52} \textcolor[rgb]{0.0,0.6,0.0}{$\uparrow$1.54}  & \textcolor[rgb]{0.0,0.6,0.0}{$\uparrow$7.74} \textcolor[rgb]{0.2,0.2,0.2}{$\downarrow$5.90} & \textcolor[rgb]{0.2,0.2,0.2}{$\downarrow$2.76} \textcolor[rgb]{0.0,0.6,0.0}{$\uparrow$2.30} & \textcolor[rgb]{0.0,0.6,0.0}{$\uparrow$0.49} \textcolor[rgb]{0.0,0.6,0.0}{$\uparrow$0.85} & \textcolor[rgb]{0.0,0.6,0.0}{$\uparrow$1.39} \textcolor[rgb]{0.0,0.6,0.0}{$\uparrow$0.67} & \textcolor[rgb]{0.0,0.6,0.0}{$\uparrow$0.17} \textcolor[rgb]{0.0,0.6,0.0}{$\uparrow$0.96}  \\ \hline
			
			DialogueGCN    \cite{ghosal2019dialoguegcn}            & 75.61   77.45       & 51.32  52.76        & \textbf{5.14  10.09}     & 30.91  32.56       & {\textbf{54.31  56.08}}       & \textbf{11.62}  \textbf{11.27}     & 42.51  44.65       & 58.74 60.55 \\
			\rowcolor{gray!30}
			DialogueGCN$^\ast$ & \textbf{78.19 77.82} & \textbf{52.27 54.11}  & 2.17 2.31 & \textbf{35.79 36.43} & {54.15 55.07} & 4.05 {2.12} & \textbf{48.31 47.22} & \textbf{60.96 60.98}\\ 
			$\Delta$ & \textcolor[rgb]{0.0,0.6,0.0}{$\uparrow$2.58} \textcolor[rgb]{0.0,0.6,0.0}{$\uparrow$0.37} & \textcolor[rgb]{0.0,0.6,0.0}{$\uparrow$0.95} \textcolor[rgb]{0.0,0.6,0.0}{$\uparrow$1.35} & \textcolor[rgb]{0.2,0.2,0.2}{$\downarrow$2.97} \textcolor[rgb]{0.2,0.2,0.2}{$\downarrow$7.78} & \textcolor[rgb]{0.0,0.6,0.0}{$\uparrow$4.88} \textcolor[rgb]{0.0,0.6,0.0}{$\uparrow$3.87} & \textcolor[rgb]{0.2,0.2,0.2}{$\downarrow$0.16} \textcolor[rgb]{0.2,0.2,0.2}{$\downarrow$1.01} & \textcolor[rgb]{0.2,0.2,0.2}{$\downarrow$7.57} \textcolor[rgb]{0.2,0.2,0.2}{$\downarrow$9.15} & \textcolor[rgb]{0.0,0.6,0.0}{$\uparrow$5.80} \textcolor[rgb]{0.0,0.6,0.0}{$\uparrow$2.57} & \textcolor[rgb]{0.0,0.6,0.0}{$\uparrow$2.22} \textcolor[rgb]{0.0,0.6,0.0}{$\uparrow$0.43}  \\ \hline
			
			MM-DFN  \cite{hu2022mm}              & \textbf{78.17 77.76}       & 52.15 50.69        & {0.00  0.00}     & 25.77  22.93       & \textbf{56.19  54.78}       & 0.00 0.00     & \textbf{48.31  47.82  }     & 60.30  59.44 \\
			\rowcolor{gray!30}
			MM-DFN$^\ast$ & {77.08 76.56} & \textbf{53.79 56.84}  & \textbf{2.07 4.11} & \textbf{38.10 31.92} & {53.63 50.53} & \textbf{4.23 7.10} & {47.99 46.08} & \textbf{60.65 59.62}\\ 
			$\Delta$ & \textcolor[rgb]{0.2,0.2,0.2}{$\downarrow$1.09} \textcolor[rgb]{0.2,0.2,0.2}{$\downarrow$1.20} & \textcolor[rgb]{0.0,0.6,0.0}{$\uparrow$1.64} \textcolor[rgb]{0.0,0.6,0.0}{$\uparrow$6.15}  & \textcolor[rgb]{0.0,0.6,0.0}{$\uparrow$2.07} \textcolor[rgb]{0.0,0.6,0.0}{$\uparrow$4.11} & \textcolor[rgb]{0.0,0.6,0.0}{$\uparrow$12.33} \textcolor[rgb]{0.0,0.6,0.0}{$\uparrow$8.99} & \textcolor[rgb]{0.2,0.2,0.2}{$\downarrow$2.56} \textcolor[rgb]{0.2,0.2,0.2}{$\downarrow$4.25} & \textcolor[rgb]{0.0,0.6,0.0}{$\uparrow$4.23} \textcolor[rgb]{0.0,0.6,0.0}{$\uparrow$7.10} & \textcolor[rgb]{0.2,0.2,0.2}{$\downarrow$0.32} \textcolor[rgb]{0.2,0.2,0.2}{$\downarrow$1.74}  & \textcolor[rgb]{0.0,0.6,0.0}{$\uparrow$0.35} \textcolor[rgb]{0.0,0.6,0.0}{$\uparrow$0.18} \\ \hline
			
			M2FNet \cite{chudasama2022m2fnet} & \textbf{68.88 67.98} & 72.76 58.66 & 5.57 3.45 & 50.09 \textbf{47.03} & {68.49 65.50} & \textbf{17.69 25.24} &  57.33 55.25&63.64 60.87\\
			\rowcolor{gray!30}
			M2FNet$^\ast$ & {68.40 67.27} & \textbf{73.15 60.37} & \textbf{9.13 11.25} & \textbf{51.77} 46.68 & \textbf{69.11 65.92} & 15.19 17.62 & \textbf{60.76 57.31} & \textbf{64.13 60.97}\\ 
			$\Delta$ & \textcolor[rgb]{0.2,0.2,0.2}{$\downarrow$4.48} \textcolor[rgb]{0.2,0.2,0.2}{$\downarrow$0.71} & \textcolor[rgb]{0.0,0.6,0.0}{$\uparrow$0.39} \textcolor[rgb]{0.0,0.6,0.0}{$\uparrow$1.71} & \textcolor[rgb]{0.0,0.6,0.0}{$\uparrow$3.56} \textcolor[rgb]{0.0,0.6,0.0}{$\uparrow$7.80} & \textcolor[rgb]{0.0,0.6,0.0}{$\uparrow$1.68} \textcolor[rgb]{0.2,0.2,0.2}{$\downarrow$0.35} & \textcolor[rgb]{0.0,0.6,0.0}{$\uparrow$0.62} \textcolor[rgb]{0.0,0.6,0.0}{$\uparrow$0.42} & \textcolor[rgb]{0.2,0.2,0.2}{$\downarrow$2.50} \textcolor[rgb]{0.2,0.2,0.2}{$\downarrow$7.62} & \textcolor[rgb]{0.0,0.6,0.0}{$\uparrow$3.43} \textcolor[rgb]{0.0,0.6,0.0}{$\uparrow$2.06} & \textcolor[rgb]{0.0,0.6,0.0}{$\uparrow$0.50} \textcolor[rgb]{0.0,0.6,0.0}{$\uparrow$0.10}  \\ \hline
			
			EmoCaps \cite{li2022emocaps} & 75.24 {\textbf{75.12}} & {63.57 63.19} & \textbf{3.45 3.03} & {\textbf{43.78 42.52}} & 58.34 57.05 & {\textbf{7.01 7.69}} & 58.79 57.54 & 63.52 62.97 \\
			\rowcolor{gray!30}
			EmoCaps$\ast$ & \textbf{76.37}  74.28 & {\textbf{66.57  64.74}}  & 3.11  2.14 & {40.17  42.35} & \textbf{63.33  62.52} & {6.21  7.05} & \textbf{59.45  60.26} & \textbf{64.93  63.88}\\ 
			$\Delta$ & \textcolor[rgb]{0.0,0.6,0.0}{$\uparrow$1.13} \textcolor[rgb]{0.2,0.2,0.2}{$\downarrow$0.84} & \textcolor[rgb]{0.0,0.6,0.0}{$\uparrow$3.00} \textcolor[rgb]{0.0,0.6,0.0}{$\uparrow$1.55} & \textcolor[rgb]{0.2,0.2,0.2}{$\downarrow$0.34} \textcolor[rgb]{0.2,0.2,0.2}{$\downarrow$0.89} & \textcolor[rgb]{0.2,0.2,0.2}{$\downarrow$3.61} \textcolor[rgb]{0.2,0.2,0.2}{$\downarrow$0.17} & \textcolor[rgb]{0.0,0.6,0.0}{$\uparrow$4.99} \textcolor[rgb]{0.0,0.6,0.0}{$\uparrow$5.47} & \textcolor[rgb]{0.2,0.2,0.2}{$\downarrow$0.80} \textcolor[rgb]{0.2,0.2,0.2}{$\downarrow$0.84} & \textcolor[rgb]{0.0,0.6,0.0}{$\uparrow$0.66} \textcolor[rgb]{0.0,0.6,0.0}{$\uparrow$2.72} & \textcolor[rgb]{0.0,0.6,0.0}{$\uparrow$1.41} \textcolor[rgb]{0.0,0.6,0.0}{$\uparrow$0.97}  \\ \hline
	\end{tabular}}
\end{table*}

\begin{table*}[htbp]
	\centering
	
	\caption{Experimental results on CMU-MOSI, and POM dataset. Methods with $\ast$ represent the method equipped with our LMAM module without any further changes. The best result in each column is in bold.}
	\label{table12}
	\renewcommand\arraystretch{1.2}
	\setlength{\tabcolsep}{12pt}{
		\begin{tabular}{c|ccccc|ccc}
			\hline
			\multirow{2}{*}{Methods}         & \multicolumn{5}{c}{CMU-MOSI}       & \multicolumn{3}{c}{POM} \\ \cline{2-9} 
			& Acc-2 ($\uparrow$) & F1 ($\uparrow$)  & MAE ($\downarrow$)  & Corr ($\uparrow$) & Acc-7 ($\uparrow$) & MAE ($\downarrow$)   & Acc ($\uparrow$)   & Corr ($\uparrow$) \\ \hline
			\multicolumn{1}{c|}{TextCNN \cite{2014Convolutional}}     & 62.2  & 62.0 & 1.42 & 0.41 & 23.3  & 0.88   & 34.0   & 0.13  \\
			\rowcolor{gray!30}
			\multicolumn{1}{c|}{TextCNN$^\ast$}     & \textbf{69.7}  & \textbf{69.6} & \textbf{1.31} & \textbf{0.49} & \textbf{24.6}  & \textbf{0.87}   & \textbf{34.0}   & \textbf{0.21}  \\ 
			$\Delta$ & \textcolor[rgb]{0.0,0.6,0.0}{$\uparrow$7.5} & \textcolor[rgb]{0.0,0.6,0.0}{$\uparrow$7.6}  &  \textcolor[rgb]{0.0,0.6,0.0}{$\uparrow$0.11} &  \textcolor[rgb]{0.0,0.6,0.0}{$\uparrow$0.08} & \textcolor[rgb]{0.0,0.6,0.0}{$\uparrow$1.3}  & \textcolor[rgb]{0.0,0.6,0.0}{$\uparrow$0.01}  & \textcolor[rgb]{0.0,0.6,0.0}{$\uparrow$0.0} & \textcolor[rgb]{0.0,0.6,0.0}{$\uparrow$0.08} \\ \hline
			
			\multicolumn{1}{c|}{bc-LSTM \cite{poria2017context}}     & 73.9  & 73.9 & 1.08 & 0.61 & 28.7  & 0.84   & 34.8   & 0.28  \\
			\rowcolor{gray!30}
			\multicolumn{1}{c|}{bc-LSTM+Att} & 75.2  & 75.2 & 1.05 & 0.62 & 34.6  & 0.82   & 35.3   & 0.30  \\
			
			\multicolumn{1}{c|}{bc-LSTM$^\ast$}     & \textbf{77.0}  & \textbf{77.1} & \textbf{0.96} & \textbf{0.64} & \textbf{34.7}  & \textbf{0.80}   & \textbf{39.1}   & \textbf{0.35}  \\ 
			$\Delta$ & \textcolor[rgb]{0.0,0.6,0.0}{$\uparrow$1.8} & \textcolor[rgb]{0.0,0.6,0.0}{$\uparrow$1.9}  & \textcolor[rgb]{0.0,0.6,0.0}{$\uparrow$0.09}  &  \textcolor[rgb]{0.0,0.6,0.0}{$\uparrow$0.02} & \textcolor[rgb]{0.0,0.6,0.0}{$\uparrow$0.1}  & \textcolor[rgb]{0.0,0.6,0.0}{$\uparrow$0.02}  & \textcolor[rgb]{0.0,0.6,0.0}{$\uparrow$3.8} & \textcolor[rgb]{0.0,0.6,0.0}{$\uparrow$0.05} \\ \hline
			
			\multicolumn{1}{c|}{DialogueRNN \cite{majumder2019dialoguernn}} & 75.7  & 75.8 & 1.02 & 0.62 & 34.6  & 0.81   & 36.2   & 0.32  \\
			\rowcolor{gray!30}
			\multicolumn{1}{c|}{DialogueRNN$^\ast$} & \textbf{77.4}  & \textbf{77.0} & \textbf{1.00} & \textbf{0.67} & \textbf{34.9}  & \textbf{0.77}   & \textbf{39.3}   & \textbf{0.37}  \\ 
			$\Delta$ & \textcolor[rgb]{0.0,0.6,0.0}{$\uparrow$1.7} & \textcolor[rgb]{0.0,0.6,0.0}{$\uparrow$1.2}  & \textcolor[rgb]{0.0,0.6,0.0}{$\uparrow$0.02}  &  \textcolor[rgb]{0.0,0.6,0.0}{$\uparrow$0.05} & \textcolor[rgb]{0.0,0.6,0.0}{$\uparrow$0.3} &\textcolor[rgb]{0.0,0.6,0.0}{$\uparrow$0.04}  & \textcolor[rgb]{0.0,0.6,0.0}{$\uparrow$3.1} & \textcolor[rgb]{0.0,0.6,0.0}{$\uparrow$0.05}\\ \hline
			
			\multicolumn{1}{c|}{DialogueGCN \cite{ghosal2019dialoguegcn}} & 74.3  & 74.0 & 1.06 & 0.61 & 29.6  & 0.83   & 35.0   & 0.29  \\
			\rowcolor{gray!30}
			\multicolumn{1}{c|}{DialogueGCN$^\ast$} & \textbf{74.9}  & \textbf{74.7} & \textbf{1.04} & \textbf{0.62} & \textbf{31.8}  & \textbf{0.79}   & \textbf{36.7}   & \textbf{0.35}  \\ 
			$\Delta$ & \textcolor[rgb]{0.0,0.6,0.0}{$\uparrow$0.6}  & \textcolor[rgb]{0.0,0.6,0.0}{$\uparrow$0.7}  &  \textcolor[rgb]{0.0,0.6,0.0}{$\uparrow$0.02} & \textcolor[rgb]{0.0,0.6,0.0}{$\uparrow$0.01}  & \textcolor[rgb]{0.0,0.6,0.0}{$\uparrow$2.2} & \textcolor[rgb]{0.0,0.6,0.0}{$\uparrow$0.04}  & \textcolor[rgb]{0.0,0.6,0.0}{$\uparrow$1.7} & \textcolor[rgb]{0.0,0.6,0.0}{$\uparrow$0.06} \\ \hline
			
			\multicolumn{1}{c|}{MM-DFN \cite{hu2022mm}}      & 76.7  & 76.7 & 0.91 & 0.68 & 33.3  & 0.79   & 42.5   & 0.41  \\
			\rowcolor{gray!30}
			\multicolumn{1}{c|}{MM-DFN$^\ast$}      & \textbf{77.1}  & \textbf{77.3} & \textbf{0.89} & \textbf{0.71} & \textbf{33.9}  & \textbf{0.76}  &\textbf{43.1}   & \textbf{0.41}  \\ 
			$\Delta$ & \textcolor[rgb]{0.0,0.6,0.0}{$\uparrow$0.4} & \textcolor[rgb]{0.0,0.6,0.0}{$\uparrow$0.6}  & \textcolor[rgb]{0.0,0.6,0.0}{$\uparrow$0.02}  & \textcolor[rgb]{0.0,0.6,0.0}{$\uparrow$0.03}  & \textcolor[rgb]{0.0,0.6,0.0}{$\uparrow$0.6} & \textcolor[rgb]{0.0,0.6,0.0}{$\uparrow$0.03}  & \textcolor[rgb]{0.0,0.6,0.0}{$\uparrow$0.6} & \textcolor[rgb]{0.0,0.6,0.0}{$\uparrow$0.0} \\ \hline
			
			\multicolumn{1}{c|}{M2FNet \cite{chudasama2022m2fnet}}      & 78.9  & 78.5 & 0.87 & 0.73 & 34.6  & 0.75   & 43.4   & 0.41  \\
			\rowcolor{gray!30}
			\multicolumn{1}{c|}{M2FNet$^\ast$}      & \textbf{79.6}  & \textbf{79.9} & \textbf{0.87} & \textbf{0.75} & \textbf{35.5}  & \textbf{0.73}   & \textbf{43.6 }  & \textbf{0.43}  \\ 
			$\Delta$ & \textcolor[rgb]{0.0,0.6,0.0}{$\uparrow$0.7} &  \textcolor[rgb]{0.0,0.6,0.0}{$\uparrow$1.4} &  \textcolor[rgb]{0.0,0.6,0.0}{$\uparrow$0} & \textcolor[rgb]{0.0,0.6,0.0}{$\uparrow$0.02}  & \textcolor[rgb]{0.0,0.6,0.0}{$\uparrow$0.9} &  \textcolor[rgb]{0.0,0.6,0.0}{$\uparrow$0.02} & \textcolor[rgb]{0.0,0.6,0.0}{$\uparrow$0.2} & \textcolor[rgb]{0.0,0.6,0.0}{$\uparrow$0.02} \\ \hline
			
			\multicolumn{1}{c|}{EmoCaps \cite{li2022emocaps}}     & 80.2  & 80.3 & 0.86 & 0.78 & 36.7  & 0.71   & 43.7   & 0.43  \\
			\rowcolor{gray!30}
			\multicolumn{1}{c|}{EmoCaps$^\ast$}                          & \textbf{82.7}  & \textbf{82.9} & \textbf{0.81} & \textbf{0.89} & \textbf{38.2}  & \textbf{0.63}   & \textbf{44.9}   & \textbf{0.47}  \\ $\Delta$ & \textcolor[rgb]{0.0,0.6,0.0}{$\uparrow$2.5} & \textcolor[rgb]{0.0,0.6,0.0}{$\uparrow$2.6}  & \textcolor[rgb]{0.0,0.6,0.0}{$\uparrow$0.05}  & \textcolor[rgb]{0.0,0.6,0.0}{$\uparrow$0.11}  & \textcolor[rgb]{0.0,0.6,0.0}{$\uparrow$1.5} & \textcolor[rgb]{0.0,0.6,0.0}{$\uparrow$0.08}  & \textcolor[rgb]{0.0,0.6,0.0}{$\uparrow$1.2} & \textcolor[rgb]{0.0,0.6,0.0}{$\uparrow$0.04} \\ \hline
	\end{tabular}}
\end{table*}

\subsection{Network Architecture of LMAM Module}
In this section, we design a network to implement the LMAM method. The overall network architecture of the LMAM module is illustrated in Figure~\ref{fig:attention}(d). From Figure~\ref{fig:attention}(d), we can observe that the LMAM module first receives three modal data as input, and then {generates} two types of vectors (i.e., query feature vector and matched feature vector) by a linear transformation layer. Subsequently, we compute the {attention} score based on these feature vectors. Finally, we {generate} the final fusion feature by conducting cross-modal feature fusion followed by a fine-tuning step.

As shown in Figure \ref{fig:attention}, there are three ways to use the proposed LMAM module, i.e., early fusion, early fusion with residual connections, and late fusion. For early fusion, we concatenate the three modalities and then input them into the LMAM module for feature fusion. For early fusion with residual connections, the concatenated features vectors of our three modalities are added to the feature vectors after feature fusion through the LMAM module. For late fusion, we extract the contextual semantic information from the model (e.g., EmoCaps) and then input it to the LMAM module for feature fusion.
It should be noted that the selection of the LMAM fusion ways depends on the baseline model itself. In the following experiment, we mainly adopt the latter two ways of fusion, i.e., early fusion with residual connections framework and late fusion.

\section{Experiments}
In this section, we conduct several experiments to verify the effectiveness of our proposed LMAM cross-modal fusion method. Specifically, the overall experimental setting is shown as follows. Firstly, we choose seven state-of-the-art DL-based approaches, including TextCNN~\cite{2014Convolutional}, bc-LSTM~\cite{poria2017context}, DialogueRNN~\cite{majumder2019dialoguernn}, DialogueGCN ~\cite{ghosal2019dialoguegcn}, MM-DFN\cite{hu2022mm}, M2FNet\cite{chudasama2022m2fnet}, and EmoCaps~\cite{li2022emocaps}, as backbones and embed the proposed LMAM fusion method into these approaches. {In particular, TextCNN uses CNN, 3D-CNN and openSMILE to extract text, video and audio features respectively and input the obtained multi-modal data into 1D-CNN to complete emotion classification.} Secondly, we compare our proposed LMAM method with four popular cross-modal fusion methods, including classical add operation and concatenate operation, and latest low-rank multi-modal fusion (LFM)~\cite{Liu2018EfficientLM}, tensor fusion network (TFN)~\cite{zadeh2017tensor}, {Dual Low-Rank Multimodal Fusion (Dual-LMF) \cite{jin-etal-2020-dual}, Low-Rank Multimodal Fusion with Self-Attention (Att-LMF) \cite{zhu2020multimodal}. and Low-Rank Multimodal Fusion with multi-modal Transformer (LMF-MulT) \cite{sahay2020low}.} Thirdly, we conduct an ablation study to verify the necessity of considering the multi-modal datasets. Finally, we apply the proposed LMAM method to other multi-modal recognition tasks. All the experiments are conducted on a PC with Intel Core i7-8700K CPU, and one GeForce RTX 3090 with 24GB memory.

\subsection{Datasets and Evaluation Metrics}
{The IEMOCAP~\cite{busso2008iemocap}, MELD~\cite{poria2018meld} CMU-MOSI \cite{zadeh2016mosi}, and POM \cite{park2014computational} datasets are widely used for conversational emotion recognition.} Therefore, {this paper selects these four benchmark datasets} to verify the effectiveness of our LMAM fusion method. The IEMOCAP dataset contains three modal data to meet the needs of multimodal research, namely video, text, and audio. The IEMOCAP dataset contains 151 dialogues and 7433 utterances of 5 actors and 5 actresses. The emotional labels of the IEMOCAP dataset were annotated by at least three experts, and they divided the labels into six categories, namely ``happy", ``neutral", ``sad", ``excited", ``angry" and ``frustrated". The MELD also includes video, text, and audio three modal data. The MELD dataset contains 13,708 utterances and 1,433 dialogues by multiple actors for 13.7 hours. The emotional labels of the MELD dataset were annotated by at least five experts, and they divided the labels into seven categories, namely ``fear", ``neutral", ``angry", ``joy", ``sadness", ``disgust” and ``surprise". {The CMU-MOSI data set is a multi-modal sentiment analysis data set, including video, audio, text and other data modalities. The CMU-MOSI dataset is annotated with an emotion label in the range [-3, 3]. The POM dataset contains 903 movie review videos, each of which is annotated with 16 speaker features. The speaker in each video is annotated with an emotion label including confidence, enthusiasm, and other characteristics.}

The IEMOCAP dataset only contains the training set and the test set, so we divide the test set into a test set and a validation set at a ratio of 8$:$2. The MELD dataset includes a training set, a test set, and a validation set. {Similarly, the CMU-MOSI and POM datasets also include a training set, a test set, and a validation set. Two popular metrics are chosen to evaluate the performance of each method on the IEMOCAP and MELD datasets, i.e., classification accuracy and $F1$ score. Five metrics are chosen to evaluate the performance of each method on the CMU-MOSI dataset, i.e., binary classification accuracy (Acc-2),  seven classification accuracy (Acc-7), mean absolute error (MAE), Pearson’s correlation (Corr), and $F1$ score. Three metrics are chosen to evaluate the performance of each method on the POM dataset, i.e., classification accuracy, MAE, and Corr.}

\begin{table}[htbp]
	\renewcommand\arraystretch{1.2}
	\setlength{\tabcolsep}{3mm}{
		\caption{The results of the equal parameters experiment on IEMOCAP and MELD datasets. The parameters of methods with $\diamond$ are incremented to be the same as methods with $\ast$. The best result in each column is in bold.}
		\label{table3}
		\begin{tabular}{l|ccccc}
			\hline
			\multirow{2}{*}{Method} & \multicolumn{1}{c}{\multirow{2}{*}{Params}} & \multicolumn{2}{c}{IEMOCAP}                           & \multicolumn{2}{c}{MELD}                              \\ \cline{3-6}
			& \multicolumn{1}{c}{}                       & \multicolumn{1}{c}{Acc.}  & \multicolumn{1}{c}{F1}    & \multicolumn{1}{c}{Acc.}  & \multicolumn{1}{c}{F1}    \\ \hline
			bc-LSTM  \cite{poria2017context}                & 0.53M                                      & 55.19                     & 54.96                     & 54.18                    & 55.89                    \\
			bc-LSTM$^\diamond$                 & 1.15M                                      & 54.37                     & 52.19                     & 53.11                     & 50.93                     \\
			bc-LSTM$^\ast$                 & 1.15M                                      & \textbf{65.55}                     & \textbf{60.29}                     & \textbf{54.97}                     & \textbf{56.85}                     \\ \hline
			MM-DFN \cite{hu2022mm}                & 2.21M                                      & 68.77                     & 68.20                     & 60.30                     & 59.44                     \\
			MM-DFN$^\diamond$                 & 2.83M                                      & 67.33                     & 67.02                     & 60.01                     & 57.65                     \\
			MM-DFN$^\ast$                 & 2.83M                                      & \textbf{69.94}                     & \textbf{69.69}                     & \textbf{60.65}                     & \textbf{59.62 }                     \\ \hline
			M2FNet  \cite{chudasama2022m2fnet}               & 8.47M                                      & 69.11                     & 69.86                     & 63.64                     & 60.87                     \\
			M2FNet$^\diamond$                 & 9.09M                                      & 69.46                     & 68.79                     & 62.49                     & 60.91                     \\
			M2FNet$^\ast$                 & 9.09M                                      & \textbf{70.31}                     & \textbf{70.07}                     & \textbf{64.13}                     & \textbf{60.97}                     \\ \hline
			EmoCaps  \cite{li2022emocaps}               & \multicolumn{1}{c}{12.74M}                 & \multicolumn{1}{c}{71.22} & \multicolumn{1}{c}{71.06} & \multicolumn{1}{c}{63.52} & \multicolumn{1}{c}{62.97} \\
			EmoCaps$^\diamond$                 & 13.36M                                     & 71.18                     & 70.56                     & 62.17                     & 62.11                     \\
			EmoCaps$^\ast$                 & 13.36M                                     & \textbf{73.67}                     & \textbf{73.01}                     & \textbf{64.93}                     & \textbf{63.88}                     \\ \hline
	\end{tabular}}
\end{table}

\subsection{Performance Verification Experiment}
To verify the {effectiveness} of our designed LMAM module, we first test our method in a plug-and-play way by directly embedding the LMAM module into seven state-of-the-art DL-based CER methods. The experimental results are shown in Table~\ref{table1}, Table~\ref{table2}, {Table~\ref{table12}}. From Table~\ref{table1}, Table~\ref{table2}, and {Table~\ref{table12}}, it can be easily seen that all the seven backbones have a significant performance improvement {on the four datasets} after using our proposed LMAM module. The performance improvement may attribute to the full interaction and fusion of different modal information in our proposed LMAM method, while the seven backbone networks only make a simple fusion of cross-modal information and thus neglect some complementary semantic information {between} different modals. Besides, we also compare the emotion recognition results of bc-LSTM$+$Att and bc-LSTM$^\ast$ (i.e.,bc-LSTM$+$LMAM), and the performance of bc-LSTM$^\ast$ is significantly better than that of the bc-LSTM$+$Att, which implies that the proposed LMAM module is better than the self attention module.

Since the above experiment embeds our proposed LMAM module into the backbones, thus it will increase the  parameter number of the backbone network. To verify that the performance improvement doesn't come from the increase of model complexity but the reasonable design of our LMAM module, we increase the parameter number of four backbones (i.e., bc-LSTM, MM-DFN, M2FNet, and EmoCaps) to the same as after embedding the LMAM module. The experimental results are shown in Table~\ref{table3}. It can be observed from Table~\ref{table3} that the performance of both bc-LSTM, MM-DFN, M2FNet, and EmoCaps methods embedded with the LMAM module are better than the bc-LSTM, MM-DFN, M2FNet, and EmoCaps models with the same parameter number, which proves that the performance improvement is not due to the increase of parameter number but is brought by our LMAM module.

{\subsection{Parameter Size and Runtime Analysis}}

{To verify that the proposed LMAM is a lightweight plug-and-play module, we analyze the changes in parameter size and running time after adding the LMAM method to the baselines on different emotion recognition datasets. In particular, we perform all experiments under the same experimental configuration to ensure the fairness of the experiments. As shown in Table \ref{tab:run}, we observe that after adding our proposed LMAM module to the baselines, the running time and model parameter size only increase slightly, but the performance of emotion recognition can be greatly improved. The experimental results show that through appropriate model optimization or architecture improvement, the computational overhead can be effectively controlled while improving the model performance. The performance improvement is attributed to the fact that the proposed LMAM can more effectively realize feature fusion, so that the model can better capture and utilize the complementary information between different modalities, thereby improving the accuracy of emotion recognition. At the same time, the increase in running time and parameter size is small, indicating that the design of the model successfully avoids excessive growth in resource requirements while retaining complexity and depth.}

\begin{table*}[htbp]
	
	\centering
	\renewcommand\arraystretch{1.2}
	\setlength{\tabcolsep}{6.5mm}{
		\caption{The results of the parameter size and running time experiment on IEMOCAP, MELD, CMU-MOSI, and POM datasets. The parameters of methods with $\diamond$ are incremented to be the same as methods with $\ast$.}
		\label{tab:run}
		\begin{tabular}{l|ccccc}
			\hline
			\multirow{2}{*}{Methods} & \multirow{2}{*}{Params} & IEMOCAP      & MELD         & CMU-MOSI     & POM          \\ \cline{3-6} 
			&                         & Running Time & Running Time & Running Time & Running Time \\ \hline
			TextCNN \cite{2014Convolutional}                  & 0.23M                   & 0.68s        & 1.15s        & 0.25s        & 0.11s        \\
			TextCNN$^\ast$                  & 0.85M                   & 1.65s        & 2.34s        & 0.32s        & 0.14s        \\ \hline
			bc-LSTM \cite{poria2017context}                  & 0.53M                   & 1.21s        & 2.17s        & 0.27s        & 0.19s        \\
			bc-LSTM+Att                  & 4.7M                    & 4.13s        & 6.63s        & 0.83s        & 0.44s        \\
			bc-LSTM$^\ast$                  & 1.15M                   & 2.18s        & 2.89s        & 0.46s        & 0.25s        \\ \hline
			DialogueRNN  \cite{majumder2019dialoguernn}            & 14.47M                  & 47.96s       & 76.92s       & 12.56s       & 7.73s        \\
			DialogueRNN$^\ast$              & 15.09M                  & 48.54s       & 78.43s       & 13.17s       & 8.42s        \\ \hline
			DialogueGCN \cite{ghosal2019dialoguegcn}             & 12.92M                  & 56.18s       & 82.34s       & 17.48s       & 9.32s        \\
			DialogueGCN$^\ast$              & 13.54M                  & 57.77s       & 88.53s       & 18.23s       & 9.74s        \\ \hline
			MM-DFN  \cite{hu2022mm}                 & 2.21M                   & 3.15s        & 6.17s        & 1.41s        & 0.71s        \\
			MM-DFN$^\ast$                   & 2.83M                   & 3.89s        & 7.04s        & 1.83s        & 0.79s        \\ \hline
			M2FNet \cite{chudasama2022m2fnet}                  & 8.47M                   & 12.48s       & 25.16s       & 3.39s        & 1.15s        \\
			M2FNet$^\ast$                   & 9.09M                   & 13.11s       & 28.33s       & 3.77s        & 1.82s        \\ \hline
			EmoCaps \cite{li2022emocaps}                 & 12.74M                  & 16.73s       & 31.07s       & 4.82s        & 2.56s        \\
			EmoCaps$^\ast$                  & 13.36M                  & 17.54s       & 34.35s       & 5.14s        & 3.06s        \\ \hline
	\end{tabular}}
\end{table*}

\begin{table*}[htbp]
	
	\caption{Statistically significant results on IEMOCAP dataset. We report P-values for each emotion category under paired t-tests.}
	\label{tab:S2}
	\renewcommand\arraystretch{1.2}
	\setlength{\tabcolsep}{15.4pt}{
		\begin{tabular}{c|cccccc}
			\hline
			\multirow{2}{*}{Methods} & \multicolumn{6}{c}{IEMOCAP}                          \\ \cline{2-7} 
			& Happy & Sad & Neutral & Angry & Excited & Frustrated \\ \hline
			TextCNN \cite{2014Convolutional}                 & 9.8 $\times 10^{-10}$     & 6.0 $\times 10^{-10}$  &  $6.1 \times 10^{-7}$       &  $1.1 \times 10^{-6}$     &   $4.8 \times 10^{-7}$      &   $ 4.2 \times 10^{-6}$         \\
			bc-LSTM \cite{poria2017context}                 &  4.8  $\times 10^{-3}$   &  $6.7 \times 10^{-4}$   &    $3.1 \times 10^{-4}$     &   $5.6 \times 10^{-5}$    &  $1.7 \times 10^{-3}$      &  $3.9 \times  10^{-3}$          \\
			bc-LSTM+Att                 &     $4.3 \times 10^{-5}$    &   $8.9 \times 10^{-8}$       &   $5.5 \times 10^{-6}$   &   $1.2\times 10^{-8}$      & 5.9 $\times 10^{-7}$   &  $3.2 \times 10^{-3}$             \\
			DialogueRNN \cite{majumder2019dialoguernn}             & 4.9 $\times 10^{-8}$     &   $3.4 \times 10^{-2}$  &      $2.9 \times 10^{-5}$   &    $2.8 \times 10^{-3}$   &      $6.4 \times 10^{-8}$   &     $1.7 \times 10^{-9}$       \\
			DialogueGCN  \cite{ghosal2019dialoguegcn}            & 3.9 $\times 10^{-7}$      & 5.2 $\times 10^{-8}$   &      $5.1 \times 10^{-7}$   &    $1.9 \times 10^{-6}$   &    $7.3 \times 10^{-3}$     & 9.4  $\times 10^{-8}$         \\
			MM-DFN \cite{hu2022mm}                  &  1.0  $\times 10^{-8}$   &  $4.5 \times 10^{-5}$   &    $5.8 \times 10^{-5}$     &  1.1  $\times 10^{-5}$   &  6.8   $\times 10^{-7}$    &  4.5  $\times 10^{-7}$        \\
			M2FNet \cite{chudasama2022m2fnet}                  & 9.3  $\times 10^{-7}$     & $3.5\times 10^{-3}$    &     $1.7 \times 10^{-4}$    &   $4.1 \times 10^{-4}$    &  5.8   $\times 10^{-7}$   & 1.1 $\times 10^{-8}$           \\
			Emocaps \cite{li2022emocaps}                 & 5.1 $\times 10^{-4}$      & 4.3 $\times 10^{-5}$    & 2.1 $\times 10^{-4}$       & 7.2 $\times 10^{-6}$     &  3.5 $\times 10^{-8}$      &  2.6   $\times 10^{-7}$       \\ \hline
	\end{tabular}}
\end{table*}

\begin{table*}[htbp]
	
	\caption{Statistically significant results on MELD dataset. We report P-values for each emotion category under paired t-tests.}
	\label{tab:S1}
	\renewcommand\arraystretch{1.2}
	\setlength{\tabcolsep}{10.5pt}{
		\begin{tabular}{c|ccccccc}
			\hline
			\multirow{2}{*}{Methods} & \multicolumn{7}{c}{MELD}                                    \\ \cline{2-8} 
			& Neutral & Surprise & Fear & Sadness & Joy & Disgust & Anger \\ \hline
			TextCNN  \cite{2014Convolutional}                &  $1.1 \times 10^{-8}$       &   $2.9 \times 10^{-3}$       &  $1.8\times 10{-6}$    &  $1.4 \times 10^{-2}$      & $7.6 \times 10^{-5}$    &  $2.1 \times 10^{-9}$       &  $8.4 \times 10^{-9}$     \\
			bc-LSTM \cite{poria2017context}                 &  $2.3 \times 10^{-4}$       &  $4.3 \times 10^{-3}$        & $2.5 \times 10^{-7}$     &   $4.6 \times 10^{-10}$      &   $7.8 \times 10^{-4}$  &  $1.1 \times 10^{-7}$       &  $5.6 \times 10^{-4}$     \\
			bc-LSTM+Att                 &   $2.4 \times 10^{-4}$      &     $5.1 \times 10^{-3}$     &   $2.5 \times 10^{-7}$   &   $1.9\times 10^{-3}$      &  $6.6 \times 10^{-4}$   &    $2.7 \times 10^{-5}$     &   $8.9 \times 10^{-4}$    \\
			DialogueRNN \cite{majumder2019dialoguernn}             &     $3.7 \times 10^{-4}$    &   $4.9 \times 10^{-3}$       &    $2.2 \times 10^{-3}$  &   $8.7 \times 10^{-8}$      &   $5.4 \times 10^{-5}$  &   $2.8 \times 10^{-3}$      &  $1.5 \times 10^{-3}$      \\
			DialogueGCN  \cite{ghosal2019dialoguegcn}             &   $1.3 \times 10^{-2}$      &   $3.4 \times 10^{-4}$       &   $1.7 \times 10^{-3}$   &   $6.7 \times 10^{-6}$      &  $3.7 \times 10^{-3}$   &     $1.2 \times 10^{-10}$    &   $5.2 \times 10^{-5}$    \\
			MM-DFN  \cite{hu2022mm}                 &  $3.7 \times 10^{-3}$       &   $7.2 \times 10^{-8}$       &  2.5 $\times 10^{-6}$   &  8.9   $\times 10^{-10}$    & 6.6 $\times 10^{-6}$   & 9.1    $\times 10^{-9}$    &  $4.7 \times 10^{-3}$     \\
			M2FNet \cite{chudasama2022m2fnet}                   &   $1.4 \times 10^{-2}$      &  6.9   $\times 10^{-4}$     &  7.6  $\times 10^{-10}$  & $3.0 \times 10^{-3}$        & 2.9  $\times 10^{-2}$  &    $9.2 \times 10^{-10}$     &  $6.1 \times 10^{-4}$     \\
			Emocaps\cite{li2022emocaps}                  &    $6.9 \times 10^{-5}$     &  1.7   $\times 10^{-3}$     & 4.4 $\times 10^{-3}$     & 4.7 $\times 10^{-2}$       & 8.4  $\times 10^{-7}$   &   $3.9 \times 10^{-2}$      &  $5.7 \times 10^{-5}$     \\ \hline
	\end{tabular}}
\end{table*}

{\subsection{Statistical Significance}}
{In this section, to further verify the significance level between the LMAM method proposed in this paper and the baseline method, we conducted a paired t-test to ensure the reliability of our experimental conclusions. The significance level results are shown in Tables \ref{tab:S1} and \ref{tab:S2}. We run the baseline model and the baseline model with the LMAM module 5 times and obtained the corresponding F1 value as an evaluation metric of significance statistics. All emotion categories on IEMOCAP and MELD datasets are statistically significant in F1 values under paired t-test ($p<0.05$).} 

\subsection{Comparison with Other Cross-modal Fusion Methods}
{In this section, to further verify the superiority of our proposed LMAM module, we also conduct an experiment to compare our LMAM module with other four typical cross-modal fusion approaches, i.e., classical add operation and concatenate operation, and latest low-rank multi-modal fusion (LFM)~\cite{Liu2018EfficientLM}, tensor fusion network (TFN)~\cite{zadeh2017tensor}, {Dual Low-Rank Multimodal Fusion (Dual-LMF) \cite{jin-etal-2020-dual}, Low-Rank Multimodal Fusion with Self-Attention (Att-LMF) \cite{zhu2020multimodal}, Low-Rank Multimodal Fusion with multi-modal Transformer (LMF-MulT) \cite{sahay2020low}, AuxFormer \cite{goncalves2022auxformer}, Cross-modality Context fusion and Semantic Refinement Network (CMCF-SRNet) \cite{zhang2023cross}, and Low-Rank Adaption (LoRA) \cite{hulora}.} The selected backbone network is EmoCaps~\cite{li2022emocaps} and the used datasets are also IEMOCAP and MELD.}

\begin{table}[H]
	\centering
	\renewcommand\arraystretch{1.2}
	\setlength{\tabcolsep}{4.8mm}{
		\caption{{The comparison results of our LMAM fusion method with other cross-modal methods. The best result is highlighted in bold. $^\diamond$ represents randomly initialized pre-training weights, and $^\ast$ using the pre-trained weights of EmoCaps.}}
		\label{table4}
		\begin{tabular}{c|cccc}
			\hline
			\multirow{2}{*}{Methods} & \multicolumn{2}{c}{IEMOCAP} & \multicolumn{2}{c}{MELD} \\ \cline{2-5}
			& Acc.       & F1              & Acc.         & F1         \\ \hline
			Add {\cite{lian2021ctnet}}                      & 71.3        & 71.0       & 64.3          & 64.0           \\
			Concatenate  \cite{cambria2018benchmarking}              & 68.9        & 68.1                & 62.6          & 61.1           \\
			TFN \cite{zadeh2017tensor} & 70.8        &  70.1               & 63.0          & 62.8           \\
			LFM \cite{Liu2018EfficientLM} & 72.3 & 71.7 & 64.1 & 62.3\\
			{Dual-LMF \cite{jin-etal-2020-dual}} & {72.6} & {72.0} & 
			{64.8} & 
			{63.2}\\
			
			{Att-LFM \cite{zhu2020multimodal}} & 
			{72.5} & 
			{71.3} & 
			{64.3} & 
			{62.9}\\
			
			{LMF-MulT \cite{sahay2020low}} & 
			{72.5} & {71.9} & {64.6} & {62.5}\\
			
			{AuxFormer \cite{goncalves2022auxformer}}  &   {71.6}   &   {72.1}   &  {63.9}   & {63.0}   \\
			
			{CMCF-SRNet \cite{zhang2023cross}}   & {72.3}    &  {71.4}   & {62.3}    & {63.3}  \\
			
			{LoRA$^\diamond$} \cite{hulora}  & {60.7}        & {58.3}                & {55.6}          & {57.5}          \\
			
			{LoRA$^\ast$} \cite{hulora}  & {73.3}        & {73.0}                & {64.4}          & {63.5}          \\
			
			{LMAM (Ours)} & \textbf{73.7}        & {\textbf{73.0}}                & {\textbf{64.9} }         & {\textbf{63.9}   }        \\ \hline
	\end{tabular}}
\end{table}

{The experimental results are recorded in Table~\ref{table4}. As shown in the Table~\ref{table4}, the LMAM method achieves the best experimental results on the IEMOCAP and MELD datasets, with Acc of 73.0\% and 65.4\%, respectively, and F1 values of 73.0\% and 64.9\%, respectively. Specifically, compared with the Add method, the Acc and F1 values of the LMAM method on the IEMOCAP dataset are increased by 1.7\% and 2.0\%, respectively, and the Acc and F1 values on the MELD dataset are increased by 1.1\% and 0.9\%, respectively. Compared with the Concatenate method, the Acc and F1 values of the LMAM method on the IEMOCAP dataset are increased by 4.1\% and 4.9\%, respectively, and the Acc and F1 values on the MELD dataset are increased by 2.8\% and 3.8\%, respectively. We think this is because the Add method and the Concatenate method do not model complementary semantic information within and between modalities. Additionally, compared with the TFN, LFM, Dual-LMF, Att-LMF, LMF-MulT, AuxFormer and CMCF-SRNet methods, the LMAM method has also achieved better performance in the accuracy and F1 value of emotion recognition, which further illustrates the superiority of our designed LMAM fusion method. Moreover, to further illustrate the effectiveness of our proposed LMAM method, we also compare it with LoRA. Specifically, we divide LoRA into two versions, one with randomly initialized pre-trained weights and the other with pre-trained weights using EmoCaps. Experimental results show that LoRA using EmoCaps' pre-trained weights performs similarly to our proposed LMAM method, but the performance of LoRA is very poor when the pre-trained weights are randomly initialized. We believe this is because the pre-trained weights of LoRA are frozen, so it is very dependent on the prior information of the pre-trained weights. Although the LMAM method we proposed randomly initializes the weights, since our method is end-to-end and the weights is learnable, it can learn weights with better generalization performance.}

\begin{table}[htbp]
	\renewcommand\arraystretch{1.2}
	\setlength{\tabcolsep}{5.5mm}{
		\caption{Experimental results of using single-modal data and multi-modal data on IEMOCAP and MELD datasets. T, A, and V represent text, audio, and video, respectively.}
		\label{table5}
		\begin{tabular}{c|cccc}
			\hline
			\multirow{2}{*}{Modality} & \multicolumn{2}{c}{IEMOCAP} & \multicolumn{2}{c}{MELD} \\ \cline{2-5}
			& Acc.       & F1              & Acc.         & F1         \\ \hline
			T                      & 70.8        & 69.2       & 63.1          & 61.8           \\
			A              & 66.0        & 65.3               & 60.4          & 59.2           \\
			V & 58.8 & 56.8 & 53.9 & 50.7\\
			{T+A}  & {72.3}  &  {71.5}  & {64.4}   & {62.7}  \\
			{T+V}  & {70.8}  & {69.6}   &  {63.5}  & {62.1}  \\
			{V+A}  & {66.7}  & {65.8}   & {61.0}   & {59.9}   \\
			T+A+V & \textbf{73.7}        & \textbf{73.0}                & \textbf{64.9}          & \textbf{63.9}           \\ \hline
	\end{tabular}}
\end{table}

\subsection{Ablation study}
\subsubsection{Necessity of multi-modal data}
To illustrate the necessity of multi-modal research, we used the EmoCaps method equipped with LMAM module as the backbone to conduct a comparative experiment of {uni-modal, bi-modal and multi-modal} on the IEMOCAP and MELD datasets. The experimental results are shown in Table \ref{table5}. We conducted a {uni-modal} experiment to utilize only one of the three modalities (i.e., text, video, and audio), {a bi-modal experiment to use any two modalities} and a multi-modal experiment to use all the three modalities. For the {uni-modal} experiments, we found that the features of the text modality performed best for emotion recognition on both datasets, followed by the features of the audio modality, and the worst performance of the features of the video modality. {In the bi-modal experiment, the emotion recognition effect of text + audio is the best, followed by the emotion recognition of text + video, and the emotion recognition of video + audio is the worst.} In the multimodal experiment, we can find that the emotion recognition effect of the combination of the three modalities is the best. Experimental results demonstrate that it is necessary to consider the multi-modal study. Furthermore, designing multi-modal feature fusion methods to improve the effect of emotion recognition is also necessary.

\subsubsection{Comparison of different embedding ways}
To compare the performance of the early fusion and early fusion with residual connections embedding ways introduced in Section 4.2, we conduct comparative experiments using the bc-LSTM, MM-DFN, M2FFNet, and EmoCaps algorithms on the IEMOCAP and MELD datasets. The experimental results are shown in Table {\ref{tab:fusion}}. As can be seen, the LMAM module with residual connections can obtain better performance compared with the early fusion without {residual} connection.

\begin{table}[htbp]
	\caption{Methods with $\ast$ represent the method equipped with our LMAM module without any further changes. Methods with $\ast$(R) represent the method equipped with our LMAM module with residual connections. The best result is highlighted in bold.}
	\label{tab:fusion}
	\renewcommand\arraystretch{1.2}
	\setlength{\tabcolsep}{4.3mm}{
		\begin{tabular}{l|cccc}
			\hline
			\multirow{2}{*}{Methods} & \multicolumn{2}{c}{IEMOCAP} & \multicolumn{2}{c}{MELD} \\ \cline{2-5}
			& Acc.          & F1          & Acc.         & F1        \\ \hline
			bc-LSTM$\ast$              &    59.49           &   59.16          &    57.69          &     55.47      \\
			bc-LSTM$\ast$(R)   & \textbf{61.77}                     & \textbf{60.49}                     & \textbf{59.56}                     & \textbf{57.49}                     \\ \hline
			MM-DFN$\ast$              &    67.93           &   67.15         &     {63.28}        &  {61.12 }        \\
			MM-DFN$\ast$(R)   & \textbf{69.82}                     & \textbf{69.68}                     &   {\textbf{ 68.02}}                   &         {\textbf{65.24} }            \\ \hline
			M2FNet$\ast$              &    68.35           &   57.96          &    67.47          &     66.59      \\
			M2FNet$\ast$(R)   & \textbf{70.27}                     & \textbf{70.07}                     & \textbf{68.34}                     & \textbf{67.25}                     \\ \hline
			EmoCaps$\ast$              &       71.49        &    71.01        &        65.03       &    64.22       \\
			EmoCaps$\ast$(R)              & \textbf{73.67}                     & \textbf{73.01}                     & \textbf{64.93}                     & \textbf{63.88}                     \\ \hline
	\end{tabular}}
\end{table}

\subsection{Complexity Analysis}
We assume that the query vectors $W^Q$, key vector  $W^K$, and value vector  $W^V$ in the self-attention mechanism have the same dimensions as the input multimodal feature $d_n$. Theoretically, the computational complexity of the LMAM method proposed in this paper is $O\left(\sum_{n=1}^M \sum_{i=1}^{r(d_n)} ({d_n^{(i)}})^3\right)$ compared to $O\left(\sum_{n=1}^M (d_n)^3\right)$ of self-attention model. {Furthermore}, we evaluate the computational complexity and computation time of LMAM and self-attention mechanism. As shown in the Table \ref{compare}, the training time and parameter of the LMAM method are much smaller than the self-attention mechanism. Here, we set the rank size to 45.

\subsection{Rank Settings}
We verified the impact of different rank parameter settings on the accuracy of emotion recognition on the IEMOCAP dataset. The experimental results are shown in the Fig \ref{fig:rank}. We observe that when $rank=45$, the training effect of the model is the best, and the training effect of the model is stable when the rank is between 30 and 55. When $rank > 45$, the training result of the model becomes unstable and the effect is poor. Therefore, better experimental results can be obtained by using lower ranks. Therefore, better experimental results can be obtained by using lower ranks.

\begin{table}
	\centering
	\caption{Comparison the training speed and parameters of self-attention mechanism and LMAM method in an epoch. Emocaps is selected as our architecture.}
	\label{compare}
	\renewcommand\arraystretch{1.2}
	\setlength{\tabcolsep}{4mm}{
		\begin{tabular}{lcc}
			\toprule
			Methods & Training time (s)  &  Parameters (M) \\ \toprule
			LMAM    &       17.8         &   0.62  \\
			Self-attention \cite{vaswani2017attention}  &   58.7     &  3.42  \\ \bottomrule
	\end{tabular}}
\end{table}

\begin{figure}
	\centering
	\includegraphics[width=0.96\linewidth]{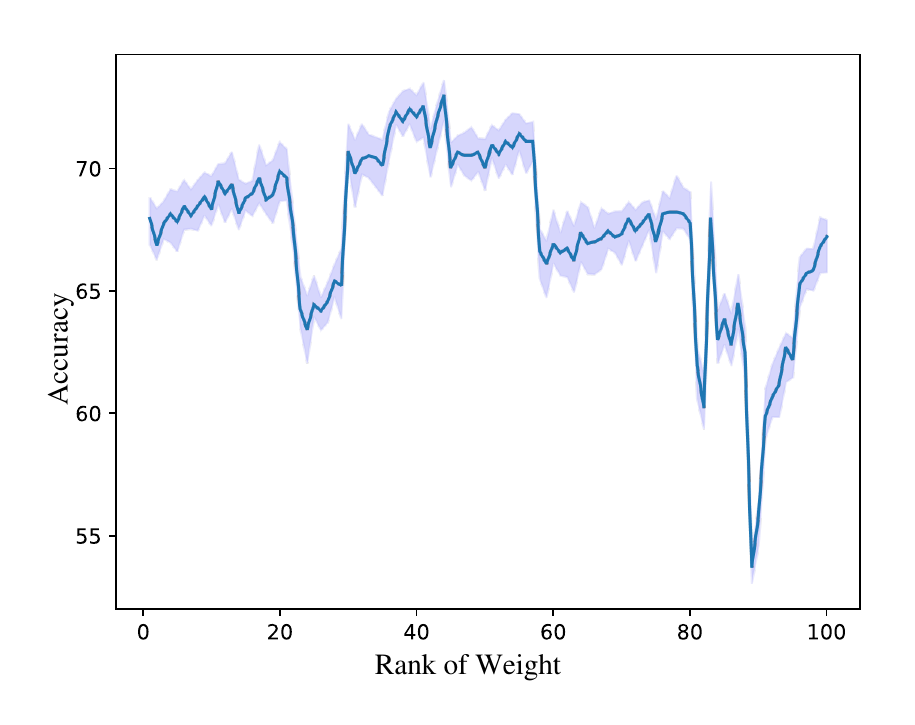}
	\caption{The impact of different rank parameter settings on the experimental accuracy. When the rank exceeds 55, the training results of the model start to be unstable.}
	\label{fig:rank}
\end{figure}

\subsection{Potential Applications}
In order to verify that our LMAM method has a potential application in other multi-modal recognition tasks, we further apply the proposed LMAM module to session recommendation and humor detection tasks. Specifically, we embed our LMAM method into dual channel hypergraph convolutional network (DHCN)~\cite{xia2021self} for session recommendation task and Contextual Memory Fusion Network (C-MFN)~\cite{zadeh2018memory} for humor detection task, respectively. The session recommendation task is conducted on the Digietica dataset\footnote{http://cikm2016.cs.iupui.edu/cikm-cup/}, and the humor detection task is carried out on the UR-FUNNY~\cite{hasan2019ur} dataset. The experimental results are illustrated in Table~\ref{table6} and Table~\ref{table7}. As shown in Table~\ref{table6} and Table~\ref{table7}, we can observe that our proposed LMAM module can improve the performance of the backbone networks in other multi-modal recognition tasks.

\begin{table}[htbp]
	\renewcommand\arraystretch{1.2}
	\setlength{\tabcolsep}{1.5mm}{
		\caption{Experimental results of DHCN method on the Digietica dataset for the session recommendation task. We use P@K (Precision) and MRR@K (Mean Reciprocal Rank) to evaluate the recommendation results. $\ast$ means the method equipped with the LMAM module. The best result is highlighted in bold.}
		\label{table6}
		\begin{tabular}{c|cccccc}
			\hline
			\multirow{2}{*}{Method} & \multicolumn{6}{c}{Digietica}                   \\ \cline{2-7}
			& P@5   & P@10  & P@20  & MRR@5 & MRR@10 & MRR@20 \\ \hline
			DHCN \cite{xia2021self}                   & 27.24 & 39.87 & 53.18 & 15.28 & 17.53  & 18.44  \\
			DHCN$^\ast$                    & \textbf{28.19} & \textbf{40.38} & \textbf{53.70} & \textbf{16.43} & \textbf{17.90}  & \textbf{18.69}  \\ \hline
	\end{tabular}}
\end{table}

\begin{table}[htbp]
	\renewcommand\arraystretch{1.2}
	\setlength{\tabcolsep}{2.9mm}{
		\caption{Experimental results of C-MFN method on the UR-FUNNY dataset for the humor detection task. C-MFN (C) means using only contextual information without punchlines. C-MFN (P) means using only punchlines with no contextual information. $\ast$ means the method equipped with the LMAM module. The best result is highlighted in bold.}
		\label{table7}
		\begin{tabular}{lccccc}
			\hline
			\multicolumn{6}{c}{UR-FUNNY}                                                 \\ \hline
			\multicolumn{1}{l|}{Modality}        & T     & A+V   & T+A   & T+V   & T+A+V \\ \hline
			\multicolumn{1}{l|}{C-MFN(P) \cite{hasan2019ur} }       & 62.85 & 53.30 & 63.28 & 63.22 & 64.47 \\
			\multicolumn{1}{l|}{C-MFN(C) \cite{hasan2019ur}}        & 57.96 & 50.23 & 57.78 & 57.99 & 58.45 \\
			\multicolumn{1}{l|}{C-MFN \cite{hasan2019ur}}          & 64.44 & 57.99 & 64.47 & 64.22 & 65.23 \\
			\multicolumn{1}{l|}{C-MFN(P)$^\ast$} & \textbf{65.43} & \textbf{59.36} & \textbf{66.04} & \textbf{66.59} & \textbf{66.87} \\
			\multicolumn{1}{l|}{C-MFN(C)$^\ast$} & \textbf{59.46} & \textbf{53.69} & \textbf{58.39} & \textbf{58.68} & \textbf{59.23} \\
			\multicolumn{1}{l|}{C-MFN$^\ast$}    & \textbf{65.66} & \textbf{59.34} & \textbf{65.68} & \textbf{64.97} & \textbf{67.00} \\ \hline
	\end{tabular}}
\end{table}

\section{Conclusion}
In this paper, we propose a novel cross-modal feature fusion method to enable better cross-modal feature fusion. {To capture the complementary emotional contextual semantic information in different modalities,} we utilize a low-rank matching attention mechanism (LMAM) to realize the interaction between multimodal features and use low-rank weights to improve efficiency. LMAM is better than the existing fusion methods while has a lower complexity. Extensive experimental results verify that LMAM can be embedded into any existing DL-based CER methods to improve their performance in a plug-and-play manner. We also mathematically prove the effectiveness of our method. Further, LMAM is a general cross-modal feature fusion method and has potential application value in other multi-modal {recognition} tasks, e.g., session recommendation and humor detection.

{Although the LMAM method can achieve better performance improvement and increase the running speed of the model, the introduction of low-rank decomposition technology may cause the model to lose contextual semantic information. Therefore, in future research work, we will study how to achieve efficient fusion of multi-modal features without losing contextual semantic information.}

\bibliographystyle{IEEEtrans}
\bibliography{refs}

\begin{thebibliography}{10}
\providecommand{\url}[1]{#1}
\csname url@samestyle\endcsname
\providecommand{\newblock}{\relax}
\providecommand{\bibinfo}[2]{#2}
\providecommand{\BIBentrySTDinterwordspacing}{\spaceskip=0pt\relax}
\providecommand{\BIBentryALTinterwordstretchfactor}{4}
\providecommand{\BIBentryALTinterwordspacing}{\spaceskip=\fontdimen2\font plus
\BIBentryALTinterwordstretchfactor\fontdimen3\font minus
  \fontdimen4\font\relax}
\providecommand{\BIBforeignlanguage}[2]{{%
\expandafter\ifx\csname l@#1\endcsname\relax
\typeout{** WARNING: IEEEtranS.bst: No hyphenation pattern has been}%
\typeout{** loaded for the language `#1'. Using the pattern for}%
\typeout{** the default language instead.}%
\else
\language=\csname l@#1\endcsname
\fi
#2}}
\providecommand{\BIBdecl}{\relax}
\BIBdecl

\bibitem{busso2008iemocap}
C.~Busso, M.~Bulut, C.-C. Lee, A.~Kazemzadeh, E.~Mower, S.~Kim, J.~N. Chang,
  S.~Lee, and S.~S. Narayanan, ``Iemocap: Interactive emotional dyadic motion
  capture database,'' \emph{Language Resources and Evaluation}, vol.~42, no.~4,
  pp. 335--359, 2008.

\bibitem{cambria2018benchmarking}
E.~Cambria, D.~Hazarika, S.~Poria, A.~Hussain, and R.~Subramanyam,
  ``Benchmarking multimodal sentiment analysis,'' in \emph{Computational
  Linguistics and Intelligent Text Processing: 18th International Conference,
  CICLing 2017, Budapest, Hungary, April 17--23, 2017, Revised Selected Papers,
  Part II 18}.\hskip 1em plus 0.5em minus 0.4em\relax Springer, 2018, pp.
  166--179.

\bibitem{chudasama2022m2fnet}
V.~Chudasama, P.~Kar, A.~Gudmalwar, N.~Shah, P.~Wasnik, and N.~Onoe, ``M2fnet:
  multi-modal fusion network for emotion recognition in conversation,'' in
  \emph{Proceedings of the IEEE/CVF Conference on Computer Vision and Pattern
  Recognition}, 2022, pp. 4652--4661.

\bibitem{dosovitskiyimage}
A.~Dosovitskiy, L.~Beyer, A.~Kolesnikov, D.~Weissenborn, X.~Zhai,
  T.~Unterthiner, M.~Dehghani, M.~Minderer, G.~Heigold, S.~Gelly \emph{et~al.},
  ``An image is worth 16x16 words: Transformers for image recognition at
  scale,'' in \emph{International Conference on Learning Representations}.

\bibitem{eyben2010opensmile}
F.~Eyben, M.~W{\"o}llmer, and B.~Schuller, ``Opensmile: the munich versatile
  and fast open-source audio feature extractor,'' in \emph{Proceedings of the
  18th ACM International Conference on Multimedia}.\hskip 1em plus 0.5em minus
  0.4em\relax ACM, 2010, pp. 1459--1462.

\bibitem{geetha2024multimodal}
A.~Geetha, T.~Mala, D.~Priyanka, and E.~Uma, ``Multimodal emotion recognition
  with deep learning: advancements, challenges, and future directions,''
  \emph{Information Fusion}, vol. 105, p. 102218, 2024.

\bibitem{ghosal2019dialoguegcn}
D.~Ghosal, N.~Majumder, S.~Poria, N.~Chhaya, and A.~Gelbukh, ``Dialoguegcn: A
  graph convolutional neural network for emotion recognition in conversation,''
  in \emph{Proceedings of the 2019 Conference on Empirical Methods in Natural
  Language Processing and the 9th International Joint Conference on Natural
  Language Processing (EMNLP-IJCNLP)}.\hskip 1em plus 0.5em minus 0.4em\relax
  ACL, 2019, pp. 154--164.

\bibitem{ghosh2022comma}
S.~Ghosh, G.~V. Singh, A.~Ekbal, and P.~Bhattacharyya, ``Comma-deer:
  Common-sense aware multimodal multitask approach for detection of emotion and
  emotional reasoning in conversations,'' in \emph{Proceedings of the 29th
  International Conference on Computational Linguistics}, 2022, pp. 6978--6990.

\bibitem{goncalves2022auxformer}
L.~Goncalves and C.~Busso, ``Auxformer: Robust approach to audiovisual emotion
  recognition,'' in \emph{ICASSP 2022-2022 IEEE International Conference on
  Acoustics, Speech and Signal Processing (ICASSP)}.\hskip 1em plus 0.5em minus
  0.4em\relax IEEE, 2022, pp. 7357--7361.

\bibitem{hasan2019ur}
M.~K. Hasan, W.~Rahman, A.~B. Zadeh, J.~Zhong, M.~I. Tanveer, L.-P. Morency,
  and M.~E. Hoque, ``Ur-funny: A multimodal language dataset for understanding
  humor,'' in \emph{Proceedings of the 2019 Conference on Empirical Methods in
  Natural Language Processing and the 9th International Joint Conference on
  Natural Language Processing (EMNLP-IJCNLP)}, 2019, pp. 2046--2056.

\bibitem{hazarika2018icon}
D.~Hazarika, S.~Poria, R.~Mihalcea, E.~Cambria, and R.~Zimmermann, ``Icon:
  Interactive conversational memory network for multimodal emotion detection,''
  in \emph{Proceedings of the 2018 Conference on Empirical Methods in Natural
  Language Processing}.\hskip 1em plus 0.5em minus 0.4em\relax ACL, 2018, pp.
  2594--2604.

\bibitem{hazarika2018conversational}
D.~Hazarika, S.~Poria, A.~Zadeh, E.~Cambria, L.-P. Morency, and R.~Zimmermann,
  ``Conversational memory network for emotion recognition in dyadic dialogue
  videos,'' in \emph{Proceedings of the conference. Association for
  Computational Linguistics.}, vol. 2018, 2018, p. 2122.

\bibitem{hu2022mm}
D.~Hu, X.~Hou, L.~Wei, L.~Jiang, and Y.~Mo, ``Mm-dfn: Multimodal dynamic fusion
  network for emotion recognition in conversations,'' in \emph{ICASSP 2022-2022
  IEEE International Conference on Acoustics, Speech and Signal Processing
  (ICASSP)}.\hskip 1em plus 0.5em minus 0.4em\relax IEEE, 2022, pp. 7037--7041.

\bibitem{hu2021dialoguecrn}
D.~Hu, L.~Wei, and X.~Huai, ``Dialoguecrn: Contextual reasoning networks for
  emotion recognition in conversations,'' in \emph{Proceedings of the 59th
  Annual Meeting of the Association for Computational Linguistics and the 11th
  International Joint Conference on Natural Language Processing (Volume 1: Long
  Papers)}, 2021, pp. 7042--7052.

\bibitem{hulora}
E.~J. Hu, P.~Wallis, Z.~Allen-Zhu, Y.~Li, S.~Wang, L.~Wang, W.~Chen
  \emph{et~al.}, ``Lora: Low-rank adaptation of large language models,'' in
  \emph{International Conference on Learning Representations}.

\bibitem{hu2021mmgcn}
J.~Hu, Y.~Liu, J.~Zhao, and Q.~Jin, ``Mmgcn: Multimodal fusion via deep graph
  convolution network for emotion recognition in conversation,'' in
  \emph{Proceedings of the 59th Annual Meeting of the Association for
  Computational Linguistics and the 11th International Joint Conference on
  Natural Language Processing (Volume 1: Long Papers)}, 2021, pp. 5666--5675.

\bibitem{jin-etal-2020-dual}
T.~Jin, S.~Huang, Y.~Li, and Z.~Zhang, ``Dual low-rank multimodal fusion,'' in
  \emph{Findings of the Association for Computational Linguistics: EMNLP 2020},
  2020, pp. 377--387.

\bibitem{2014Convolutional}
Y.~Kim, ``Convolutional neural networks for sentence classification,'' in
  \emph{Proceedings of the 2014 Conference on Empirical Methods in Natural
  Language Processing ({EMNLP})}.\hskip 1em plus 0.5em minus 0.4em\relax ACL,
  2014, pp. 1746--1751.

\bibitem{kolda2009tensor}
T.~G. Kolda and B.~W. Bader, ``Tensor decompositions and applications,''
  \emph{SIAM review}, vol.~51, no.~3, pp. 455--500, 2009.

\bibitem{li2024cfn}
J.~Li, X.~Wang, Y.~Liu, and Z.~Zeng, ``Cfn-esa: A cross-modal fusion network
  with emotion-shift awareness for dialogue emotion recognition,'' \emph{IEEE
  Transactions on Affective Computing}, 2024.

\bibitem{li2023graphmft}
J.~Li, X.~Wang, G.~Lv, and Z.~Zeng, ``Graphmft: A graph network based
  multimodal fusion technique for emotion recognition in conversation,''
  \emph{Neurocomputing}, vol. 550, p. 126427, 2023.

\bibitem{li2022contrast}
S.~Li, H.~Yan, and X.~Qiu, ``Contrast and generation make bart a good dialogue
  emotion recognizer,'' in \emph{Proceedings of the AAAI Conference on
  Artificial Intelligence}, vol.~36, no.~10, 2022, pp. 11\,002--11\,010.

\bibitem{li2022emocaps}
Z.~Li, F.~Tang, M.~Zhao, and Y.~Zhu, ``Emocaps: Emotion capsule based model for
  conversational emotion recognition,'' in \emph{Findings of the Association
  for Computational Linguistics: ACL 2022}, 2022, pp. 1610--1618.

\bibitem{lian2021ctnet}
Z.~Lian, B.~Liu, and J.~Tao, ``Ctnet: Conversational transformer network for
  emotion recognition,'' \emph{IEEE/ACM Transactions on Audio, Speech, and
  Language Processing}, vol.~29, pp. 985--1000, 2021.

\bibitem{liu2022social}
H.~Liu, K.~Li, J.~Fan, C.~Yan, T.~Qin, and Q.~Zheng, ``Social image--text
  sentiment classification with cross-modal consistency and knowledge
  distillation,'' \emph{IEEE Transactions on Affective Computing}, vol.~14,
  no.~4, pp. 3332--3344, 2022.

\bibitem{Liu2018EfficientLM}
Z.~Liu, Y.~Shen, V.~B. Lakshminarasimhan, P.~P. Liang, A.~B. Zadeh, and L.-P.
  Morency, ``Efficient low-rank multimodal fusion with modality-specific
  factors,'' in \emph{Proceedings of the 56th Annual Meeting of the Association
  for Computational Linguistics (Volume 1: Long Papers)}.\hskip 1em plus 0.5em
  minus 0.4em\relax ACL, 2018, pp. 2247--2256.

\bibitem{10109845}
H.~Ma, J.~Wang, H.~Lin, B.~Zhang, Y.~Zhang, and B.~Xu, ``A transformer-based
  model with self-distillation for multimodal emotion recognition in
  conversations,'' \emph{IEEE Transactions on Multimedia}, pp. 1--13, 2023.

\bibitem{majumder2019dialoguernn}
N.~Majumder, S.~Poria, D.~Hazarika, R.~Mihalcea, A.~Gelbukh, and E.~Cambria,
  ``Dialoguernn: An attentive rnn for emotion detection in conversations,'' in
  \emph{Proceedings of the AAAI Conference on Artificial Intelligence},
  vol.~33, no.~01.\hskip 1em plus 0.5em minus 0.4em\relax AAAI, 2019, pp.
  6818--6825.

\bibitem{mohammad2022ethics}
S.~M. Mohammad, ``Ethics sheet for automatic emotion recognition and sentiment
  analysis,'' \emph{Computational Linguistics}, vol.~48, no.~2, pp. 239--278,
  2022.

\bibitem{park2014computational}
S.~Park, H.~S. Shim, M.~Chatterjee, K.~Sagae, and L.-P. Morency,
  ``Computational analysis of persuasiveness in social multimedia: A novel
  dataset and multimodal prediction approach,'' in \emph{Proceedings of the
  16th International Conference on Multimodal Interaction}, 2014, pp. 50--57.

\bibitem{poria2017context}
S.~Poria, E.~Cambria, D.~Hazarika, N.~Majumder, A.~Zadeh, and L.-P. Morency,
  ``Context-dependent sentiment analysis in user-generated videos,'' in
  \emph{Proceedings of the 55th Annual Meeting of the Association for
  Computational Linguistics (volume 1: Long papers)}.\hskip 1em plus 0.5em
  minus 0.4em\relax ACL, 2017, pp. 873--883.

\bibitem{poria2018meld}
S.~Poria, D.~Hazarika, N.~Majumder, G.~Naik, E.~Cambria, and R.~Mihalcea,
  ``Meld: A multimodal multi-party dataset for emotion recognition in
  conversations,'' in \emph{Proceedings of the 57th Annual Meeting of the
  Association for Computational Linguistics}.\hskip 1em plus 0.5em minus
  0.4em\relax ACL, 2019, pp. 527--536.

\bibitem{ren2021lr}
M.~Ren, X.~Huang, W.~Li, D.~Song, and W.~Nie, ``Lr-gcn: Latent relation-aware
  graph convolutional network for conversational emotion recognition,''
  \emph{IEEE Transactions on Multimedia}, pp. 1--1, 2021.

\bibitem{sahay2020low}
S.~Sahay, E.~Okur, S.~H. Kumar, and L.~Nachman, ``Low rank fusion based
  transformers for multimodal sequences,'' in \emph{Second Grand-Challenge and
  Workshop on Multimodal Language (Challenge-HML)}, 2020, pp. 29--34.

\bibitem{schuster2022bert}
C.~Schuster and S.~Hegelich, ``From bert‘s point of view: Revealing the
  prevailing contextual differences,'' in \emph{Findings of the Association for
  Computational Linguistics: ACL 2022}, 2022, pp. 1120--1138.

\bibitem{shahin2022novel}
I.~Shahin, N.~Hindawi, A.~B. Nassif, A.~Alhudhaif, and K.~Polat, ``Novel
  dual-channel long short-term memory compressed capsule networks for emotion
  recognition,'' \emph{Expert Systems with Applications}, vol. 188, p. 116080,
  2022.

\bibitem{sheng2020summarize}
D.~Sheng, D.~Wang, Y.~Shen, H.~Zheng, and H.~Liu, ``Summarize before aggregate:
  a global-to-local heterogeneous graph inference network for conversational
  emotion recognition,'' in \emph{Proceedings of the 28th International
  Conference on Computational Linguistics}.\hskip 1em plus 0.5em minus
  0.4em\relax ICCL, 2020, pp. 4153--4163.

\bibitem{shou2022conversational}
Y.~Shou, T.~Meng, W.~Ai, S.~Yang, and K.~Li, ``Conversational emotion
  recognition studies based on graph convolutional neural networks and a
  dependent syntactic analysis,'' \emph{Neurocomputing}, vol. 501, pp.
  629--639, 2022.

\bibitem{su2024dynamic}
Y.~Su, Y.~Wei, W.~Nie, S.~Zhao, and A.~Liu, ``Dynamic causal disentanglement
  model for dialogue emotion detection,'' \emph{IEEE Transactions on Affective
  Computing}, 2024.

\bibitem{tu2022exploration}
G.~Tu, J.~Wen, H.~Liu, S.~Chen, L.~Zheng, and D.~Jiang, ``Exploration meets
  exploitation: Multitask learning for emotion recognition based on discrete
  and dimensional models,'' \emph{Knowledge-Based Systems}, vol. 235, p.
  107598, 2022.

\bibitem{vaswani2017attention}
A.~Vaswani, N.~Shazeer, N.~Parmar, J.~Uszkoreit, L.~Jones, A.~N. Gomez,
  {\L}.~Kaiser, and I.~Polosukhin, ``Attention is all you need,''
  \emph{Advances in neural information processing systems}, vol.~30, 2017.

\bibitem{xia2021self}
X.~Xia, H.~Yin, J.~Yu, Q.~Wang, L.~Cui, and X.~Zhang, ``Self-supervised
  hypergraph convolutional networks for session-based recommendation,'' in
  \emph{Proceedings of the AAAI conference on artificial intelligence},
  vol.~35, no.~5, 2021, pp. 4503--4511.

\bibitem{9128015}
S.~Xing, S.~Mai, and H.~Hu, ``Adapted dynamic memory network for emotion
  recognition in conversation,'' \emph{IEEE Transactions on Affective
  Computing}, vol.~13, no.~3, pp. 1426--1439, 2022.

\bibitem{yang2022hybrid}
L.~Yang, Y.~Shen, Y.~Mao, and L.~Cai, ``Hybrid curriculum learning for emotion
  recognition in conversation,'' in \emph{Proceedings of the AAAI Conference on
  Artificial Intelligence}, vol.~36, no.~10, 2022, pp. 11\,595--11\,603.

\bibitem{zadeh2017tensor}
A.~Zadeh, M.~Chen, S.~Poria, E.~Cambria, and L.-P. Morency, ``Tensor fusion
  network for multimodal sentiment analysis,'' in \emph{Proceedings of the 2017
  Conference on Empirical Methods in Natural Language Processing}, 2017, pp.
  1103--1114.

\bibitem{zadeh2018memory}
A.~Zadeh, P.~P. Liang, N.~Mazumder, S.~Poria, E.~Cambria, and L.-P. Morency,
  ``Memory fusion network for multi-view sequential learning,'' in
  \emph{Proceedings of the AAAI conference on artificial intelligence},
  vol.~32, no.~1, 2018.

\bibitem{zadeh2016mosi}
A.~Zadeh, R.~Zellers, E.~Pincus, and L.-P. Morency, ``Mosi: multimodal corpus
  of sentiment intensity and subjectivity analysis in online opinion videos,''
  \emph{arXiv preprint arXiv:1606.06259}, 2016.

\bibitem{zhang2023cross}
X.~Zhang and Y.~Li, ``A cross-modality context fusion and semantic refinement
  network for emotion recognition in conversation,'' in \emph{Proceedings of
  the 61st Annual Meeting of the Association for Computational Linguistics
  (Volume 1: Long Papers)}, 2023, pp. 13\,099--13\,110.

\bibitem{zhang2024cross}
Y.~Zhang, H.~Liu, D.~Wang, D.~Zhang, T.~Lou, Q.~Zheng, and C.~Quek,
  ``Cross-modal credibility modelling for eeg-based multimodal emotion
  recognition,'' \emph{Journal of Neural Engineering}, vol.~21, no.~2, p.
  026040, 2024.

\bibitem{zhu2020multimodal}
H.~Zhu, Z.~Wang, Y.~Shi, Y.~Hua, G.~Xu, and L.~Deng, ``Multimodal fusion method
  based on self-attention mechanism,'' \emph{Wireless Communications and Mobile
  Computing}, vol. 2020, pp. 1--8, 2020.

\end{thebibliography}

\begin{IEEEbiography}[{\includegraphics[width=1in,height=1.25in,clip,keepaspectratio]{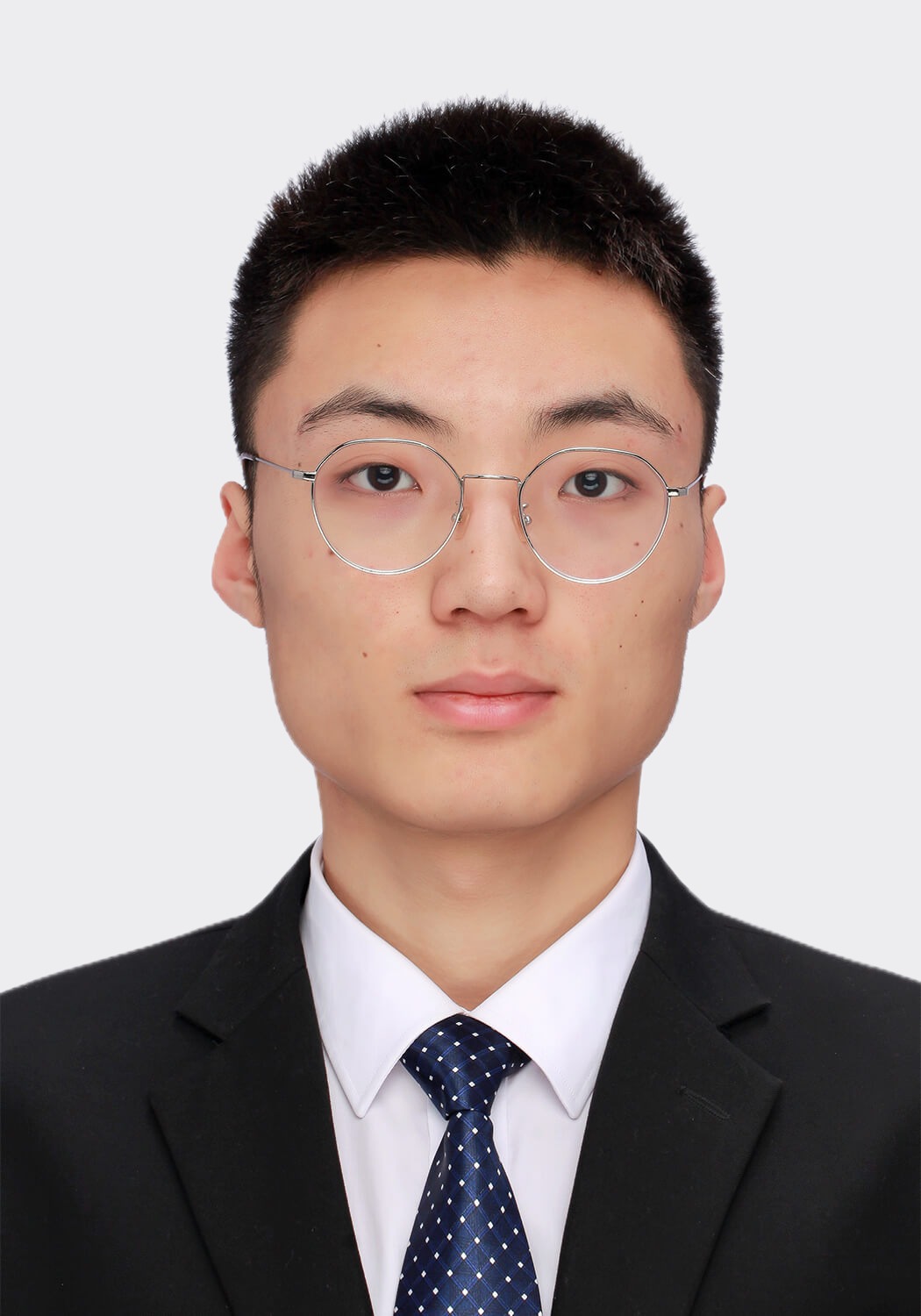}}]{Yuntao Shou} is currently pursuing the M.S. degree in the School of Computer Science and Technology, Xi'an Jiaotong University, Xi'an. His research interests include graph learning and emotion recognition. (Email: shouyuntao@stu.xjtu.edu.cn)
\end{IEEEbiography}

\begin{IEEEbiography}[{\includegraphics[width=1in,height=1.25in,clip,keepaspectratio]{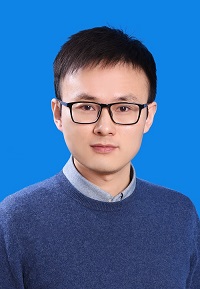}}]{Huan Liu}  received the B.Sc. and Ph.D. degrees from Xi’an
	Jiaotong University, Xi’an, China, in 2013 and 2020, respec
	tively. He is currently an Associate Professor with the School
	of Computer Science and Technology, Xi’an Jiaotong Uni
	versity. His research interests include affective computing,
	machine learning, and deep learning. (Email: huanliu@xjtu.edu.cn)
\end{IEEEbiography}

\begin{IEEEbiography}[{\includegraphics[width=1in,height=1.25in,clip,keepaspectratio]{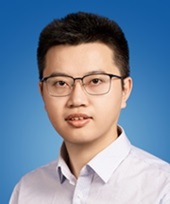}}]{Xiangyong Cao}  received the B.Sc. and Ph.D. degrees from
	Xi’an Jiaotong University, Xi’an, China, in 2012 and 2018,
	respectively. From 2016 to 2017, he was a Visiting Scholar
	with Columbia University, New York, NY, USA. He is currently an Associate Professor with the School of Computer
	Science and Technology, Xi’an Jiaotong University. His
	research interests include statistical modeling and image
	processing. (Email: caoxiangyong@mail.xjtu.edu.cn)
\end{IEEEbiography}

\begin{IEEEbiography}[{\includegraphics[width=1in,height=1.25in,clip,keepaspectratio]{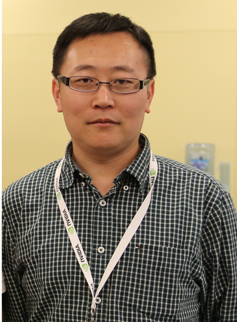}}]{Deyu Meng} received the B.Sc., M.Sc., and Ph.D. degrees
	from Xi’an Jiaotong University, Xi’an, China, in 2001, 2004,
	and 2008, respectively. From 2012 to 2014, he took his
	two-year sabbatical leave with Carnegie Mellon University,
	Pittsburgh, PA, USA. He is currently a Professor with
	the School of Mathematics and Statistics, Xi’an Jiaotong
	University, and an Adjunct Professor with the Faculty of
	Information Technology, The Macau University of Science
	and Technology, Macau, China. His research interests in
	clude model-based deep learning, variational networks, and
	meta-learning (Email: dymeng@mail.xjtu.edu.cn)
\end{IEEEbiography}

\begin{IEEEbiography}[{\includegraphics[width=1in,height=1.25in,clip,keepaspectratio]{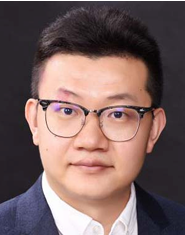}}]{Bo Dong}  received the Ph.D. de
	gree in computer science and technology from Xi’an
	Jiaotong University, Xi’an, China, in 2014. From
	2014 to 2017, he was Postdoc Researcher of control
	science and engineering, Xi’an Jiaotong University,
	where he is currently a Professor with the School of
	Distance Education. His main research interests are
	data mining, and pattern recognition. (Email: dong.bo@mail.xjtu.edu.cn)
\end{IEEEbiography}

\vfill

\end{document}